\documentclass{article}

 \usepackage[preprint]{neurips_2026}

\usepackage[utf8]{inputenc} 
\usepackage[T1]{fontenc}    
\usepackage{hyperref}       
\usepackage{url}            
\usepackage{booktabs}       
\usepackage{amsfonts}       
\usepackage{nicefrac}       
\usepackage{microtype}      
\usepackage{xcolor}         
\usepackage{xspace}  
\usepackage{graphicx}
\usepackage{booktabs}
\usepackage{adjustbox}
\usepackage{multicol}
\usepackage{multirow}
\usepackage{hyphenat}
\usepackage{subcaption}
\usepackage{bm}
\usepackage{amsmath}
\usepackage{makecell}
\usepackage{pifont}
\usepackage{lipsum}
\usepackage{bbding}
\usepackage{wasysym}
\usepackage{amssymb}
\usepackage{wrapfig}

\usepackage[dvipsnames]{xcolor} 

\newcommand{\eg}{e.g.\xspace}

\newcommand{\datasetname}{WildCity\xspace}
\newcommand{\modelname}{Wild3R\xspace}

\newcommand{\numscenes}{200\xspace}
\newcommand{\numlights}{170\xspace}
\newcommand{\numimages}{337,500\xspace}
\newcommand{\Fref}[1]{Fig.~\ref{#1}}
\newcommand{\Tref}[1]{Table~\ref{#1}}

\title{\modelname: Feed-Forward 3D Gaussian Splatting\\ from Unconstrained Sparse Photo Collections}

\author{
  \vspace{-25pt}\\
  \textbf{Yuto Furutani$^{*}$,\quad Takashi Otonari$^{*}$,\quad Kaede Shiohara\thanks{Co-first authors.},\quad Toshihiko Yamasaki
  }\vspace{3pt} \\
  The University of Tokyo\vspace{3pt} \\
  \texttt{\small \{furutani, otonari, shiohara, yamasaki\}@cvm.t.u-tokyo.ac.jp}\vspace{3pt} \\ 
  Project page:~\, \url{https://furuschool.github.io/wild3r-page}
  \vspace{-4pt} \\
}

\begin{document}

\maketitle
\begin{figure}[htb]
\centering
\includegraphics[width=.95\linewidth]{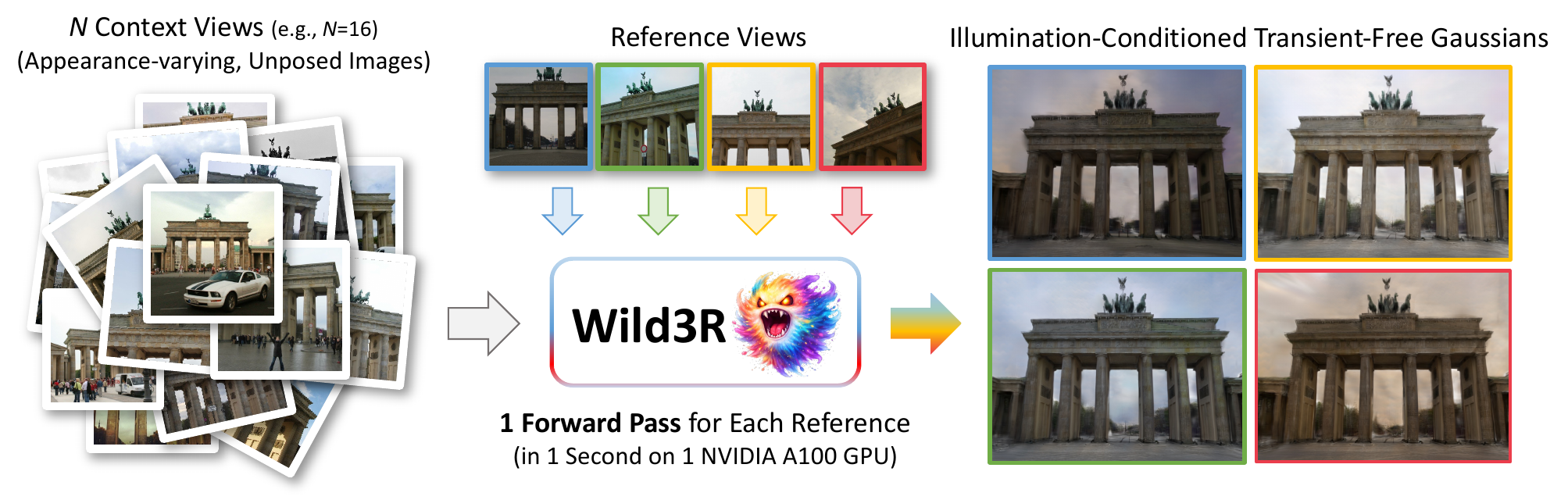}
\caption{Given unconstrained photo collections and reference views, \modelname reconstructs 3D scenes in the reference appearances without transient objects.}
\label{fig:teaser}
\end{figure}
\begin{abstract}
Feed-forward 3D Gaussian Splatting (3DGS) removes the need for time-consuming per-scene optimization required by traditional 3DGS.
However, existing feed-forward approaches struggle with real-world photo collections that include diverse lighting conditions and transient objects.
In this paper, we present \modelname, a feed-forward approach for unconstrained sparse photo collections.
The main bottleneck is the lack of training data that provides multiple viewpoints, a variety of illuminations, and transient variations necessary for learning robust scene representations. 
To address this, we introduce the \datasetname dataset, which comprises \numscenes scenes, \numlights lighting conditions, and transient objects, resulting in \numimages images in total. 
By leveraging the dataset, our model learns appearance consistency across viewpoints conditioned on reference views, while removing transient content.
Extensive experiments demonstrate that our method outperforms existing feed-forward approaches and achieves results competitive with prior per-scene optimization-based methods.
\end{abstract}

\section{Introduction}
\label{sec:intro} 
Real-world image collections exhibit uncontrolled illumination, exposure differences, seasonal variation, and transient objects such as moving objects or occluders, requiring methods to separate persistent scene structure from observation-dependent appearance.
Early neural scene representations such as Neural Radiance Fields (NeRF)~\cite{nerf} achieved high-fidelity novel view synthesis but relied on controlled captures. 
NeRF in the Wild (NeRF-W)~\cite{martinbrualla2020nerfw} addressed this limitation by modeling per-image appearance and transient objects for consistent reconstruction from Internet photo collections. 

3D Gaussian Splatting (3DGS)~\cite{kerbl3Dgaussians} introduced explicit radiance primitives that enable real-time rendering while preserving photorealistic quality. 
Subsequent works further extended it to unconstrained photo collections~\cite{swag,zhang2024GS-W,wild-gs,kulhanek2024wildgaussians,li2025asymgs,wildsplatting}.
Despite their fast rendering, these methods still rely on iterative optimization for each scene. 
Recently, feed-forward Gaussian splatting~\cite{splatt3r,flare,jiang2025anysplat,yonosplat,longlrm} has addressed this bottleneck by predicting a complete Gaussian representation directly from input images in a single forward pass, leveraging large-scale training on multi-view image datasets~\cite{co3dv2,blendedmvs,dl3dv,megadepth,kubric,wildrgb,scannet,hypersim,mapillary,habitat,replica,mvssynth,pointodyssey,virtualkitti,aria,objaverse}.

However, existing feed-forward approaches struggle to generalize to unconstrained sparse photo collections captured in the wild, where viewpoints are sparse, lighting varies significantly, and transient or view-dependent content frequently appears.
In such settings, they often produce duplicated geometry, inconsistent density, and unstable appearance across viewpoints. 
We attribute this limitation to two fundamental challenges: 1) the lack of large-scale multi-view training datasets covering diverse observation conditions, and 2) the absence of mechanisms to enforce appearance consistency and transient-free geometry across views.

To address these challenges, we propose \textbf{\modelname}, a feed-forward 3DGS method for unconstrained sparse photo collections.
As shown in \Fref{fig:teaser}, \modelname takes sparse (\eg, 16) unstructured views and a reference view to infer appearance-consistent Gaussians conditioned on the reference within one second on a single NVIDIA A100 GPU. 
Crucially, \modelname achieves this without requiring a complex, specialized architecture. 
It is realized by fine-tuning the existing feed-forward model~\cite{jiang2025anysplat}, demonstrating that the primary bottleneck lies in the training data.

\begin{figure}[t]
\centering
\begin{minipage}{0.38\linewidth}{
\includegraphics[width=0.98\linewidth]{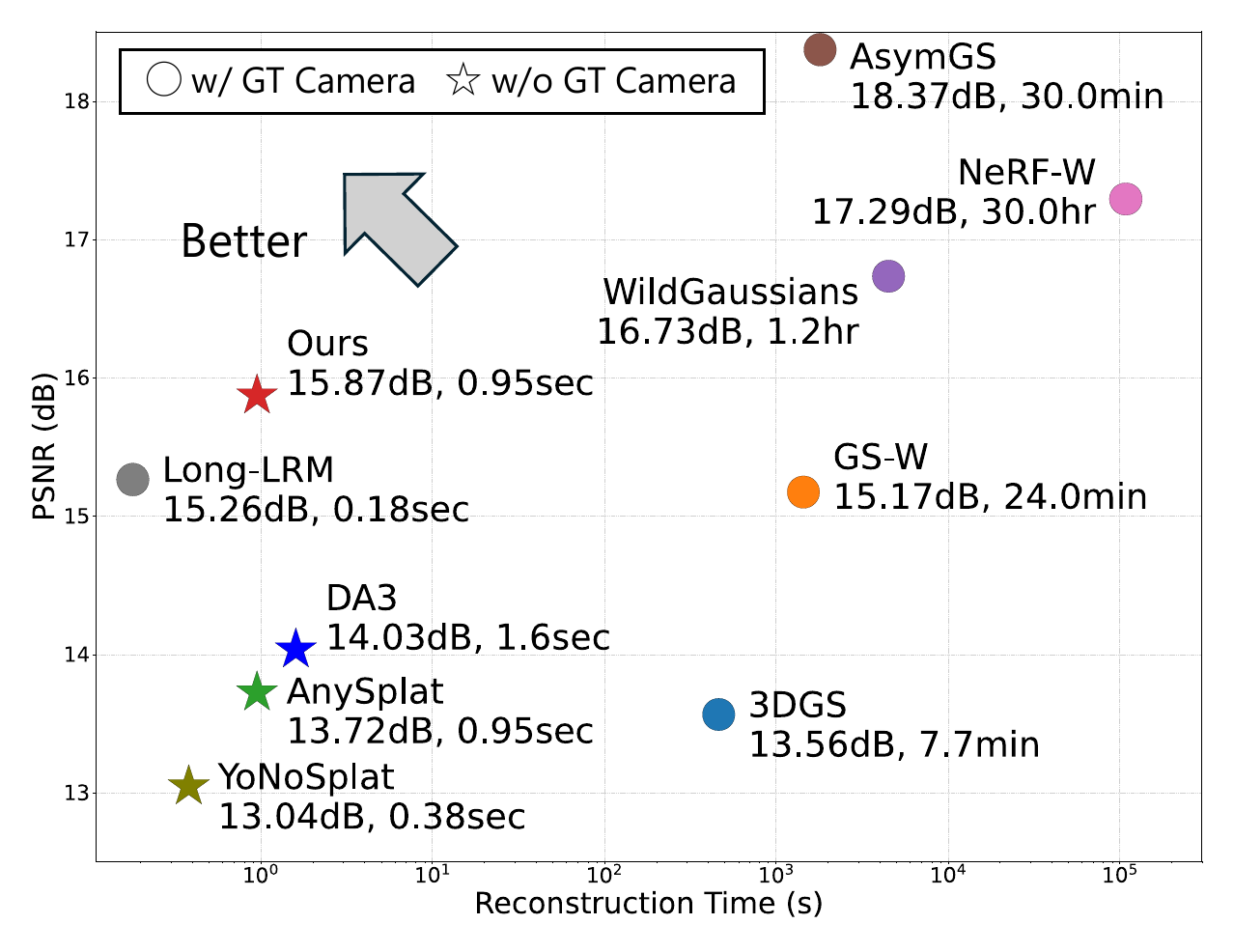}
}

\end{minipage}
~\hfill
\begin{minipage}{0.6\linewidth}{
\caption{\textbf{PSNR vs. Reconstruction Time.} 
The horizontal axis represents the reconstruction time in log scale to generate 3D Gaussians from input images. The vertical axis indicates the corresponding PSNR of each model on the Photo Tourism dataset~\cite{phototourism} using 16 context views. Star markers denote methods that operate without ground truth camera parameters, while circle markers represent those that require them.
}
\label{fig:time_vs_psnr}
}
\end{minipage}
\vspace{-10pt}
\end{figure}

\modelname is trained on our newly introduced large-scale synthetic dataset, named \textbf{\datasetname}, which comprises multi-view, multi-lighting images covering \numscenes scenes, \numlights HDRI maps, and diverse transient objects.
Leveraging the \datasetname dataset, \modelname learns to infer illumination-consistent Gaussians conditioned on the reference view while removing transient objects.
Our experiments demonstrate that our model outperforms feed-forward baselines across standard benchmarks while being competitive with per-scene optimization-based methods that require camera calibration and point cloud initialization, as shown in \Fref{fig:time_vs_psnr}.
Extensive experiments on in-the-wild benchmarks demonstrate that our approach significantly outperforms prior feed-forward methods and achieves performance competitive with optimization-based approaches, while being orders of magnitude faster at inference time.

\section{Related Work}
\label{sec:related}
\subsection{3D Reconstruction from Unconstrained Photo Collections}
In real-world image collections, illumination variations and transient objects are inevitable.
To address these challenges, several NeRF- and 3DGS-based approaches~\cite{chen2022hallucinated,sun2022neural,k-planes,kassab2025refinedfields,yang2023cross,swag,zhang2024GS-W,wild-gs,li2025asymgs,wildsplatting} have proposed to disentangle static scene content from transient objects and varying illumination using per-image appearance embeddings, as pioneered by NeRF-W~\cite{martinbrualla2020nerfw}.
However, these methods often rely on per-scene, computationally expensive optimization (\eg, NeRF-W takes dozens of hours per scene) and typically require known camera parameters, leaving practical challenges for real-world environments.

\subsection{Feed-Forward 3D Gaussian Splatting}
To address the inefficiency of per-scene optimization methods such as NeRF~\cite{nerf} and 3DGS~\cite{kerbl3Dgaussians}, recent studies~\cite{splatt3r,flare,jiang2025anysplat,depthanything3,yonosplat,offthegrid,anchorsplat} focus on feed-forward reconstruction learned from large-scale datasets.
These approaches are typically based on geometry foundation models such as DUSt3R~\cite{dust3r} and VGGT~\cite{wang2025vggt}, and extend them to feed-forward 3DGS by employing an additional head to infer per-pixel Gaussians.
Despite the impressive reconstruction ability, these models often fail to handle unconstrained photo collections since they assume a static scene captured at a single point in time.
In contrast to the existing feed-forward 3DGS methods, our model generates appearance-consistent 3D Gaussians from images captured under varying conditions over time.

\begin{table}[t]
    \centering 
    \begin{adjustbox}{width=1.0\linewidth}
    \setlength{\tabcolsep}{4pt}
    \begin{tabular}{lccccccc} 
    \toprule
    Dataset& Level &\#Images & \#Scenes & \#Illuminations & \#Illum. per View & \#Views per Scene & Transient Objects \\  
    \midrule
    DTU~\cite{dtu} & Object & 47,467 & 124 & 7 & 7 & 49 or 64 & {\color{red}\ding{55}} \\
    ReNe~\cite{Toschi_2023_CVPR} & Object & 40,000 & 20 & 40 & 40 & 50 & {\color{red}\ding{55}} \\
    OpenIllumination~\cite{openillumination}& Object & 108,096 & 64 & 155 & 155 & 72 & {\color{red}\ding{55}} \\
    OpenSubstance~\cite{opensubstance}& Object & 2,409,750 & 187 & 1,637 & 53 & 270 & {\color{red}\ding{55}} \\
    Photo Tourism~\cite{phototourism}& Scene & 29,796 & 26 & - & 1 & 75--3,765 & {\color{Green}\ding{51}} \\
    NeRF-OSR~\cite{rudnev2022nerfosr}& Scene & 3,240 & 8 & - & 1 & 331--493 & {\color{Green}\ding{51}} \\
    LightCity~\cite{lightcity}& Scene & 5,023 & 5 & 211 & 1 or 2 & 165--896 & {\color{red}\ding{55}} \\
    \textbf{\datasetname (Ours)} & Scene  & \numimages & \numscenes & \numlights & 30 & 50 & {\color{Green}\ding{51}} \\
    \bottomrule
    \end{tabular}
    \end{adjustbox}
    \vspace{1.5mm}
  \caption{\textbf{Comparative Dataset Properties: \datasetname Dataset and Existing Datasets.}}
  \label{tb:dataset}
  \vspace{-2em}
\end{table}

\subsection{In-the-Wild Datasets}
The Photo Tourism dataset~\cite{phototourism} and NeRF-OSR dataset~\cite{rudnev2022nerfosr} serve as standard benchmarks to evaluate NeRF- and 3DGS-based models in in-the-wild scenarios. 
LightCity~\cite{lightcity} provides multi-illumination renderings of outdoor urban scenes generated using SceneCity~\cite{scenecity} and Blender’s Cycles engine, but offers only a few images per viewpoint and a few scenes, and does not include transient objects.
Although DTU~\cite{dtu}, ReNe~\cite{Toschi_2023_CVPR}, OpenIllumination~\cite{openillumination}, and OpenSubstance~\cite{opensubstance} offer multiple images per camera view primarily for relighting, they lack transient objects and are limited to object-centric, simplistic scenes. 
In contrast to these existing datasets, our dataset provides sufficient multi-view images, scenes, illuminations, and transient objects to train feed-forward models as summarized in \Tref{tb:dataset}.

\section{\datasetname Dataset}
\label{sec:dataset}

\begin{figure}[t]
  \centering
  \begin{adjustbox}{width=1.0\linewidth}
  \includegraphics{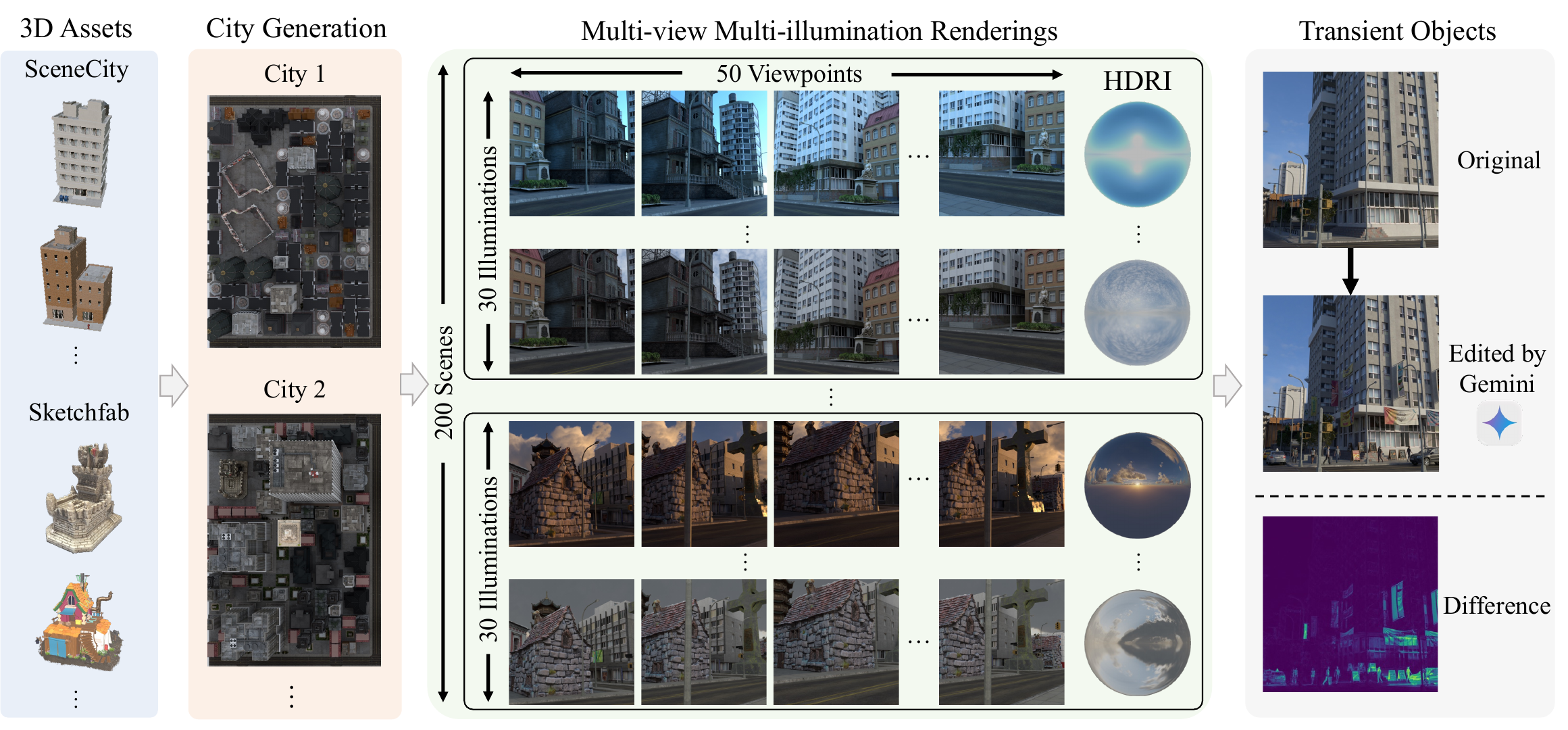}
  \end{adjustbox}
  \caption{\textbf{\datasetname Dataset Creation Pipeline.} (i) First, we collect 3D assets from the SceneCity, a Blender add-on, and Sketchfab (Sec.~\ref{subsec:data_collection}). (ii) Then, we generate scenes using these assets (Sec.~\ref{subsec:scene_generation}). (iii) Next, we render the scenes from multiple viewpoints and multiple lighting conditions using the HDRI maps (Sec.~\ref{subsec:rendering}). (iv) Finally, we add transient objects to the rendered images using Gemini (Sec.~\ref{subsec:adding transient objects}).
  } 
  \label{fig:pipeline}
\end{figure}

Our \datasetname dataset is designed to train feed-forward 3DGS models for 3D reconstruction from unconstrained photo collections. 
An overview of the dataset creation pipeline is shown in \Fref{fig:pipeline}. The dataset creation process consists of four main stages: (i) Data Collection (Sec.~\ref{subsec:data_collection}), utilizing the SceneCity Blender add-on and 3D assets from Sketchfab; (ii) Scene Generation (Sec.~\ref{subsec:scene_generation}), which involves determining the center of the scenes; (iii) Image Rendering (Sec.~\ref{subsec:rendering}); and (iv) Adding Transient Objects with Gemini~\cite{gemini} (Sec.~\ref{subsec:adding transient objects}).

\subsection{Data Collection}
\label{subsec:data_collection}

\noindent\textbf{3D Assets.}
To construct the \datasetname dataset, we used the SceneCity Blender add-on, which automatically generates 3D cities. 
However, the add-on provides only 11 building types, which significantly restricts the geometric and textural diversity necessary for training robust 3D reconstruction models.
Therefore, we augmented the default SceneCity building set with over 130 publicly available 3D models from Sketchfab (provided primarily in GLB format with permissive licenses).
These collected assets include not only typical residential and commercial buildings but also unique structures such as temples, shrines, towers, and giant Buddha statues.

\noindent\textbf{Multi-illumination.}
We utilized \numlights HDRI maps available from LightCity~\cite{lightcity} to achieve diverse lighting conditions, excluding extremely dark nighttime maps to ensure sufficient scene visibility.
During dataset generation, we randomly sampled 30 HDRI maps per scene from this collection and rendered multi-view images under each selected illumination condition.

\subsection{Scene Generation}
\label{subsec:scene_generation}
\noindent\textbf{City Generation.}
We divided the downloaded Sketchfab models into nine distinct subsets. 
By integrating each subset with the default SceneCity buildings, we constructed a total of nine unique virtual cities. 
Because SceneCity requires assets to be aligned to a 10-meter grid, we uniformly scaled and spatially normalized the downloaded GLB assets during integration to prevent severe scale distortions. 
The cities were then procedurally generated using SceneCity's built-in placement algorithms, where we assigned a higher placement probability to the Sketchfab assets than to the default buildings.

\noindent\textbf{Scene Designation.}
Next, we defined the spatial layout and focal points for each scene. 
We manually selected target buildings facing roads with an unobstructed view on the opposite side.
This careful selection prevents camera-geometry collisions and occlusions during the subsequent multi-view rendering phase. 
We sampled 20 to 25 such target locations per city, resulting in a total of \numscenes distinct scenes across the nine cities.

\subsection{Rendering}
\label{subsec:rendering}
\noindent\textbf{Camera Setup.}
To create the dataset, we sampled 50 views per scene. 
Cameras were randomly distributed within a fan-shaped region extending from the scene center, at heights between 1 and 2 meters to simulate a standard human eye level.
The camera poses were randomized such that their optical axes intersected points within a rectangular region around the scene center.
This strategy ensures sufficient overlapping regions even with sparse views, while simultaneously replicating typical human photo collections. 
Furthermore, the field of view (FoV) of each camera was uniformly sampled between $40^\circ$ and $100^\circ$ to capture the diverse focal lengths found in unconstrained photo collections. 
To prevent distant cameras from capturing empty background regions beyond the modeled ground plane, we dynamically reduced the FoV to crop out these invalid areas.

\noindent\textbf{Rendering Engine.}
We render images from the sampled camera viewpoints using Blender Cycles, a physically based renderer (PBR) capable of generating photorealistic images. 
All images are at a resolution of $512 \times 512$ pixels with 512 samples per pixel.
PBR models light transport using bidirectional scattering distribution function shaders, including diffuse reflection, glossy reflection, and transmission. 
The rendered image intensity $I(\mathbf{x})$ at pixel location $\mathbf{x}$ is modeled as
\[
I(\mathbf{x}) = I_D(\mathbf{x}) + I_G(\mathbf{x}) + I_T(\mathbf{x}) + I_B(\mathbf{x}) + I_E(\mathbf{x}),
\]
where $I_D$, $I_G$, $I_T$, $I_B$, and $I_E$ denote the diffuse, glossy, transmission, background, and emission components, respectively.
Both the diffuse and glossy components are further decoupled into direct and indirect lighting terms:
\[
I_{(D,G)}(\mathbf{x}) = \alpha_{(D,G)}(\mathbf{x})\left( I_{(D,G),\mathrm{dir}}(\mathbf{x}) + I_{(D,G),\mathrm{indir}}(\mathbf{x}) \right),
\]
where $\alpha_{(D,G)}(\mathbf{x})$ denotes the (diffuse, glossy) color and $I_{(D,G),\mathrm{dir}}(\mathbf{x}), I_{(D,G),\mathrm{indir}}(\mathbf{x})$ denote the direct and indirect lighting paths.

\subsection{Adding Transient Objects}
\label{subsec:adding transient objects}
Unconstrained photo collections inevitably contain transient objects, such as moving vehicles, pedestrians, and other entities that temporarily occlude the scene.
However, reproducing the diversity by explicitly placing objects in 3D scenes is difficult. 
We instead use a text-driven image editing model~\cite{gemini} to introduce realistic and diverse objects rather than inserting them as 3D assets; note that these transient objects are not necessarily consistent across views.
For example, we input the prompt ``Please add a person, a car, a construction warning sign, fabric banners on some buildings, while maintaining the geometry and lightness/darkness.'' into the model.
We apply this 2D augmentation to 12.5\% of the rendered views, yielding a total of 37,500 transient-augmented images.
More details can be found in the Appendix.

\begin{figure}[t]
  \centering
  \begin{adjustbox}{width=1.0\linewidth}
  \includegraphics{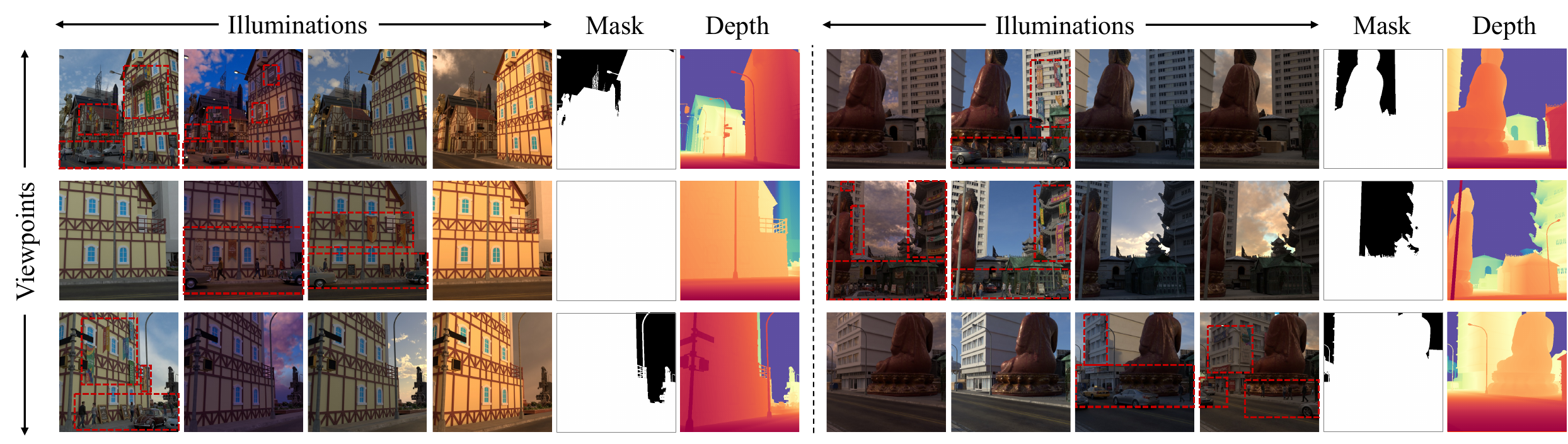}
  \end{adjustbox}
  \caption{\textbf{Scene Examples from \datasetname Dataset.} \datasetname dataset contains images of various 3D assets captured under different viewpoints and illuminations. The added transient objects are highlighted with red dashed boxes. We also provide corresponding depth maps and masks indicating the sky regions.} 
  \vspace{-5pt}
  \label{fig:wildcity_example}
  
\end{figure}

\section{\modelname}
\label{sec:model}
We aim to reconstruct 3D scenes from appearance-varying input views without per-scene optimization.
To this end, we propose \textbf{\modelname}, a feed-forward 3DGS model trained to enforce appearance consistency and transient-free geometry across views.
Our network builds upon a camera-free feed-forward 3DGS model~\cite{jiang2025anysplat} (Sec.~\ref{sec:preliminary}) and is fine-tuned on our \datasetname dataset with our newly introduced learning objectives (Sec.~\ref{sec:learning}).
This minimal extension requires no structural modifications to the base model, thereby preserving its fast inference speed and architectural simplicity.

\subsection{Preliminary}
\label{sec:preliminary}
\noindent\textbf{3D Gaussian Splatting (3DGS).}
3DGS~\cite{kerbl3Dgaussians} represents a 3D scene as a collection of anisotropic Gaussian primitives.
Each Gaussian models a localized volumetric element parameterized by
\(
(\bm{\mu}_g, \alpha_g, \bm{q}_g, \bm{s}_g, \bm{c}_g),
\)
where $\bm{\mu}_g \in \mathbb{R}^3$ denotes the 3D center position,
$\alpha_g \in \mathbb{R}$ represents opacity,
$\bm{q}_g \in \mathbb{R}^4$ is the rotation encoded as a quaternion,
$\bm{s}_g \in \mathbb{R}^3$ defines the anisotropic scaling, and
$\bm{c}_g$ represents view-dependent color parameterized using spherical harmonics coefficients.

\noindent\textbf{AnySplat.}
We base our network architecture on AnySplat~\cite{jiang2025anysplat}, which is built upon VGGT~\cite{wang2025vggt} and extends it to feed-forward 3DGS by adding an extra head to predict per-pixel Gaussian primitives.
Given a set of $N$ input views $\{I_n\}_{n=1}^N$ where $I_n \in \mathbb{R}^{H\times W\times 3}$, the model predicts the parameters of $G$ 3D Gaussians $\{(\bm{\mu}_g, \alpha_g, \bm{q}_g, \bm{s}_g, \bm{c}_g)\}_{g=1}^{G}$, along with the corresponding camera parameters and depth maps.
By convention, the camera of the first frame defines the world coordinate system for the reconstructed scene.

\begin{figure}[t]
  \centering
  \begin{adjustbox}{width=0.98\linewidth}
  \includegraphics{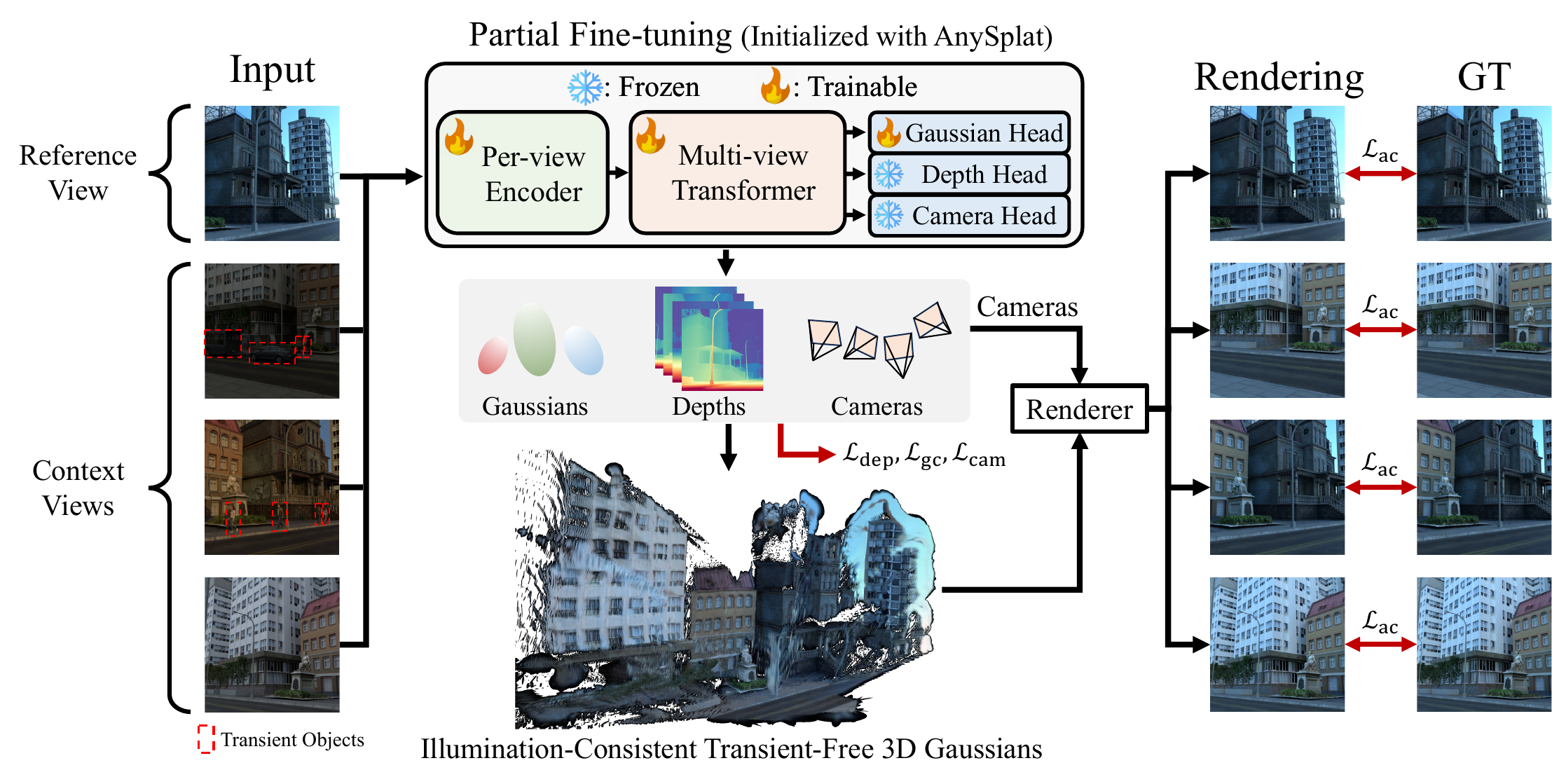}
  \end{adjustbox}
  \caption{\textbf{Learning Appearance Consistency and Transient-free Geometry.} 
  During training, \modelname takes multi-view images with random view-dependent illuminations and transient objects, and is constrained to predict transient-free 3D Gaussian primitives, depth maps, and camera poses. The illumination of the reconstructed scene is conditioned on the reference view. 
  }
  \label{fig:learning}
  \vspace{-1.5em}
\end{figure}

\subsection{Learning Appearance Consistency and Transient-free Geometry}
\label{sec:learning}
\noindent\textbf{Appearance Consistency.}
We start by taking the first frame as an appearance reference to condition the scene reconstruction.
As shown in \Fref{fig:learning}, for each training iteration, we sample a set of views
$\{ I_n^{(L_n)} \}_{n=1}^{N}$
from the training dataset, where $L_n$ denotes the lighting condition associated with the $n$-th observation.
We then apply transient augmentation to obtain $\tilde{I}_n^{(L_n)} = \tau(I_n^{(L_n)})$, where $\tau(\cdot)$ represents a stochastic augmentation pipeline. 
This pipeline applies transient object insertion with a probability of $p=0.5$.
The first augmented view $\tilde{I}_1^{(L_1)}$ serves as the reference appearance.

The feed-forward network predicts a Gaussian scene
$\mathcal{G} = h_g\left(f\left(\{\tilde{I}_n^{(L_n)}\}_{n=1}^{N}\right)\right)$,
where $f$ is the transformer-based backbone and $h_g$ is the Gaussian head.
The predicted Gaussian scene $\mathcal{G}$ is then rendered from viewpoint $n$ to produce $\hat{I}_n$.
We supervise each rendered view using a target image $I_n^{(L_1)}$ conditioned on the reference illumination $L_1$, which encourages our model to reconstruct entire scenes with the reference illumination without transient objects.
The appearance consistency loss is
\begin{equation}
\mathcal{L}_{\mathrm{ac}} = \sum_{n=1}^{N}
\left(
\lVert \hat{I}_n - I_n^{(L_1)} \rVert_2^2
+
\lambda_{\mathrm{lpips}} \, \mathcal{L}_{\mathrm{lpips}}(\hat{I}_n, I_n^{(L_1)})
\right),
\end{equation}
where $\mathcal{L}_{\mathrm{lpips}}$ denotes the LPIPS loss~\cite{lpips} and $\lambda_{\mathrm{lpips}}$ denotes a loss weight.

\noindent\textbf{Transient-free Geometry.}
Unlike previous feed-forward models such as AnySplat~\cite{jiang2025anysplat}, we supervise the predicted depth maps with the transient-free ground truth depth maps rather than the depth maps of input images. 
We denote our depth loss as $\mathcal{L}_{\mathrm{dep}}$.

\noindent\textbf{Overall Objective.}
We train our model with the following objective:
\begin{equation}
\mathcal{L} = \mathcal{L}_{\mathrm{ac}} 
+ \lambda_{\mathrm{dep}}\mathcal{L}_{\mathrm{dep}}
+ \lambda_{\mathrm{gc}} \mathcal{L}_{\mathrm{gc}}
+ \lambda_{\mathrm{cam}}\mathcal{L}_{\mathrm{cam}},
\end{equation}
where the geometric consistency loss $\mathcal{L}_{\mathrm{gc}}$ and the camera loss $\mathcal{L}_{\mathrm{cam}}$ follow AnySplat~\cite{jiang2025anysplat}.
$\lambda_{\mathrm{gc}}$, $\lambda_{\mathrm{dep}}$, and $\lambda_{\mathrm{cam}}$ are scalar coefficients to balance losses.

\section{Experiments}
\label{sec:experiments}

\subsection{Implementation Details}
\label{subsec:implementation-details}
We initialize our model using the pretrained weights of AnySplat~\cite{jiang2025anysplat}. 
We freeze the depth and camera heads while other parts of our model are updated as shown in \Fref{fig:learning}.
In total, our architecture comprises approximately 940M learnable parameters.
Following AnySplat, we train the model using the AdamW optimizer for 30K iterations with a cosine learning rate schedule, a peak learning rate of $2 \times 10^{-4}$, and a 1K-iteration warmup phase. 
Training takes approximately one day on a single NVIDIA A100 (80GB) GPU.
At each iteration, we randomly select a scene from the \datasetname dataset and sample $N \in [2, 24]$ camera views to construct a training batch.
A reference view and a reference lighting condition are then determined. 
Whenever a transient-augmented version of a selected view is available, we utilize it with a probability of $p=0.5$.
For each context view, we independently sample from images containing transient objects with a probability of $p=0.5$, and subsequently sample a lighting condition at random.
For training, we set $\lambda_{\mathrm{lpips}}, \lambda_{\mathrm{gc}}, \lambda_{\mathrm{dep}},$ and $\lambda_{\mathrm{cam}}$ to $0.05, 0.02, 0.2,$ and $2.0$, respectively.
Please refer to the Appendix for more details.

\subsection{Experimental Setup}
\noindent\textbf{Baselines.}
We compare our \modelname with prior approaches designed for unconstrained photo collections, including NeRF-W~\cite{martinbrualla2020nerfw}, GS-W~\cite{zhang2024GS-W}, WildGaussians~\cite{kulhanek2024wildgaussians}, and AsymGS~\cite{li2025asymgs}.
We further compare against 3D Gaussian Splatting (3DGS)~\cite{kerbl3Dgaussians}, camera-known feed-forward method Long-LRM~\cite{longlrm}, and camera-free feed-forward 3DGS methods such as AnySplat~\cite{jiang2025anysplat}, YoNoSplat~\cite{yonosplat}, and Depth Anything 3 (DA3, DA3NESTED-GIANT-LARGE)~\cite{depthanything3}.

\noindent\textbf{Dataset and Evaluation Protocol.}
We evaluate the models on the Photo Tourism dataset~\cite{phototourism} using three commonly used scenes: Brandenburg Gate, Sacre Coeur, and Trevi Fountain. 
Additionally, evaluations on the NeRF-OSR dataset~\cite{rudnev2022nerfosr} are provided in the Appendix.
For each scene, we construct context sets with $N\in\{4,16,64\}$ images sampled once from the official training split.
For evaluation, we use all images from the official test split.
Following the NeRF-W evaluation protocol~\cite{martinbrualla2020nerfw}, each selected test image is vertically split into two halves.
One half serves as the reference view for appearance adaptation, while the other half is used as the evaluation target.
More details can be found in the Appendix.

\noindent\textbf{Known Cameras and Point Clouds.}
Several previous methods~\cite{martinbrualla2020nerfw,kerbl3Dgaussians,zhang2024GS-W,li2025asymgs,kulhanek2024wildgaussians,longlrm} need camera parameters and/or point cloud initialization for reconstruction.
Therefore, we used the official camera parameters and initialized the point clouds corresponding to the training images.
Note that our method reconstructs scenes only from images without known camera parameters and point clouds.

\noindent\textbf{Metrics.}
We report PSNR, SSIM~\cite{ssim}, and LPIPS~\cite{lpips} following the convention. 

\begin{table}[t]
    \centering 
    \begin{adjustbox}{width=1.0\linewidth}
    \begin{tabular}{lcccccccccccc} 
    \toprule
    \multirow{2}{*}{Method} & \multirow{2}{*}{Unknown}& \multirow{2}{*}{Unknown} & \multirow{2}{*}{Reconstruction} & \multicolumn{3}{c}{4 Context Views}  & \multicolumn{3}{c}{16 Context Views} & \multicolumn{3}{c}{64 Context Views}\\
    \cmidrule(lr){5-7}\cmidrule(lr){8-10}  \cmidrule(lr){11-13}
    &Camera & Point Cloud& Time* & PSNR$\uparrow$ &  SSIM$\uparrow$ & LPIPS$\downarrow$& PSNR$\uparrow$ &  SSIM$\uparrow$ & LPIPS$\downarrow$& PSNR$\uparrow$ &  SSIM$\uparrow$ & LPIPS$\downarrow$\\  
    \midrule
    \multicolumn{2}{l}{\textit{Optimization-based}} \\
    NeRF-W~\cite{martinbrualla2020nerfw} &{\color{red}\ding{55}}&{\color{red}\ding{55}}& 30h &12.08 & 0.382 & 0.738 & 17.29 & 0.530 & 0.570 & 21.10 & 0.671 & 0.449\\
    3DGS~\cite{kerbl3Dgaussians} &{\color{red}\ding{55}}&{\color{red}\ding{55}} & 7.7m & 12.42 & 0.378 & 0.612 & 13.56 & 0.437 & 0.560 & 15.20 & 0.564 & 0.462\\
    GS-W~\cite{zhang2024GS-W} &{\color{red}\ding{55}}&{\color{red}\ding{55}} & 24m & 12.72 & 0.397 & 0.582 & 15.17 & 0.501 & 0.504 & 17.91 & 0.634 & 0.404\\
    WildGaussians~\cite{kulhanek2024wildgaussians} &{\color{red}\ding{55}}& {\color{red}\ding{55}} & 1.2h & 14.00 & 0.428 & 0.592 & 16.73 & 0.551 & 0.524 & 20.20 & 0.695 & 0.395\\
    AsymGS~\cite{li2025asymgs} &{\color{red}\ding{55}}&{\color{red}\ding{55}} & 30m& \textbf{15.93} & \textbf{0.506} & \textbf{0.574} & \textbf{18.37} & \textbf{0.607} & \textbf{0.463} & \textbf{21.24} & \textbf{0.718} & \textbf{0.340}\\
    \midrule
    \multicolumn{2}{l}{\textit{Camera-known Feed-forward}}\\
    Long-LRM~\cite{longlrm} & {\color{red}\ding{55}}&{\color{Green}\ding{51}} & 0.18s & 11.25 & 0.415 & 0.650 & 15.26 & 0.486 & 0.569 & 15.90 & 0.525 & 0.535\\
    \midrule
    \multicolumn{2}{l}{\textit{Camera-free Feed-forward}}\\
    AnySplat~\cite{jiang2025anysplat} &{\color{Green}\ding{51}}&{\color{Green}\ding{51}} & 0.95s& 11.25 & 0.320 & 0.593 & 13.72 & 0.377 & 0.546 & 14.88 & 0.417 & 0.512\\
    YoNoSplat~\cite{yonosplat} & {\color{Green}\ding{51}}&{\color{Green}\ding{51}} & 0.38s & 12.47 & \textbf{0.397} & 0.666 & 13.04 & 0.403 & 0.640 & 13.25 & 0.412 & 0.640\\
    DA3~\cite{depthanything3} & {\color{Green}\ding{51}}&{\color{Green}\ding{51}} & 1.6s & \textbf{13.35} & 0.394 & 0.622 & 14.03 & 0.420 & 0.586 & 14.25 & 0.434 & 0.582
\\
    \modelname (Ours) & {\color{Green}\ding{51}}&{\color{Green}\ding{51}} &0.95s & 13.04 & 0.370 & \textbf{0.556} & \textbf{15.87} & \textbf{0.435} & \textbf{0.506} & \textbf{16.29} & \textbf{0.458} & \textbf{0.477}\\
    \bottomrule
    \end{tabular}
    \end{adjustbox}
    \vspace{1.5mm}
  \caption{\textbf{Comparison with Previous Methods on the Photo Tourism Dataset.} The best results for both optimization-based and camera-free feed-forward models are highlighted in \textbf{bold}. \\ *Reconstruction time is recorded with 16 context views on a single NVIDIA A100 GPU.}
  \label{tb:phototourism}
  \vspace{-1em}
\end{table}

\subsection{Evaluation on the Photo Tourism Dataset}
\noindent\textbf{Quantitative Comparison.}
The results on the Photo Tourism dataset~\cite{phototourism} are reported in \Tref{tb:phototourism}. 
As the number of context views increases, optimization-based methods show a consistent tendency to improve in performance.
On the other hand, these methods require optimization during reconstruction, and they additionally assume access to extra information such as camera parameters and point clouds, which imposes constraints on their applicability. 
In contrast, feed-forward methods eliminate the need for test-time optimization, though they generally benefit less from an increased number of views.
Compared to camera-free feed-forward approaches, our method achieves superior performance across most metrics.
Notably, even under the 4-context-view setting, \modelname attains favorable LPIPS, demonstrating its robust capability to synthesize perceptually high-quality reconstructions from highly limited observations.

\begin{figure*}[t]
\centering
\includegraphics[width=\linewidth]{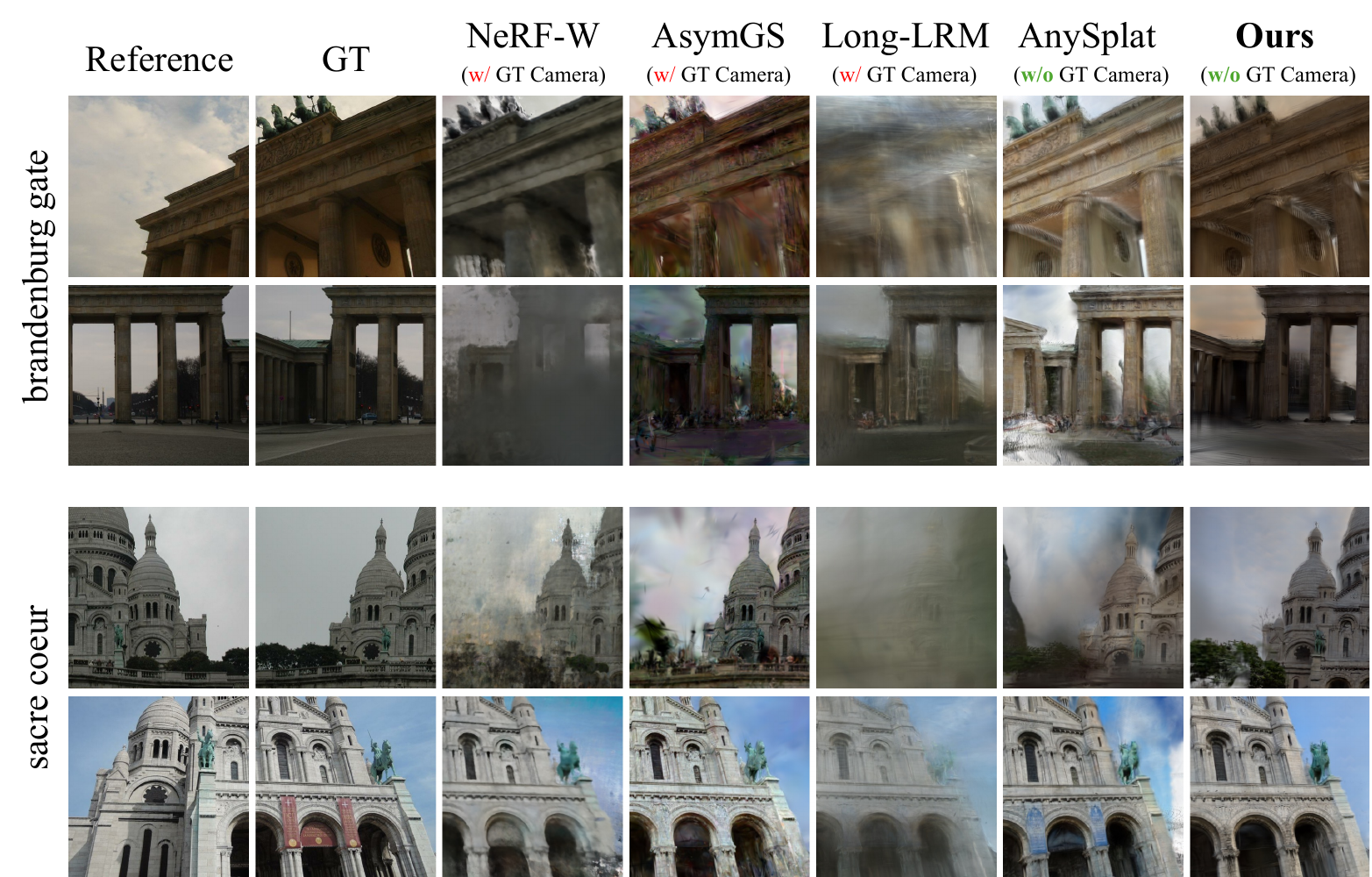}
    \caption{\textbf{Qualitative Comparison with 16 Context Views on the Photo Tourism Dataset.}}
\label{fig:example}
\vspace{-8mm}
\end{figure*}

\noindent\textbf{Qualitative Comparison.}
Qualitative comparisons on the Photo Tourism dataset~\cite{phototourism} are shown in \Fref{fig:example}. 
Our approach avoids the artifacts that existing optimization-based approaches such as NeRF-W~\cite{martinbrualla2020nerfw} and AsymGS~\cite{li2025asymgs} tend to suffer from, while also representing a lighting environment closer to the target compared to Long-LRM~\cite{longlrm} and AnySplat~\cite{jiang2025anysplat}.

\subsection{Ablation Study}
\Tref{tb:ablation} presents modeling and data ablations of our proposed method on the Photo Tourism dataset~\cite{phototourism}.
In the modeling comparison, we consider two variants: the pretrained AnySplat (a) and a fine-tuned AnySplat on \datasetname dataset using the same learning strategy as the original AnySplat (b).
As shown in the table, the vanilla fine-tuned model (b) generally improves the metrics and tends to outperform the base AnySplat~\cite{jiang2025anysplat} (a). 
However, even with these improvements, many settings still fall short of our full model, suggesting that vanilla fine-tuning alone leaves a non-trivial gap.

\begin{figure}[t]
\centering

\begin{subfigure}{0.49\linewidth}
    \centering
    \includegraphics[width=\linewidth]{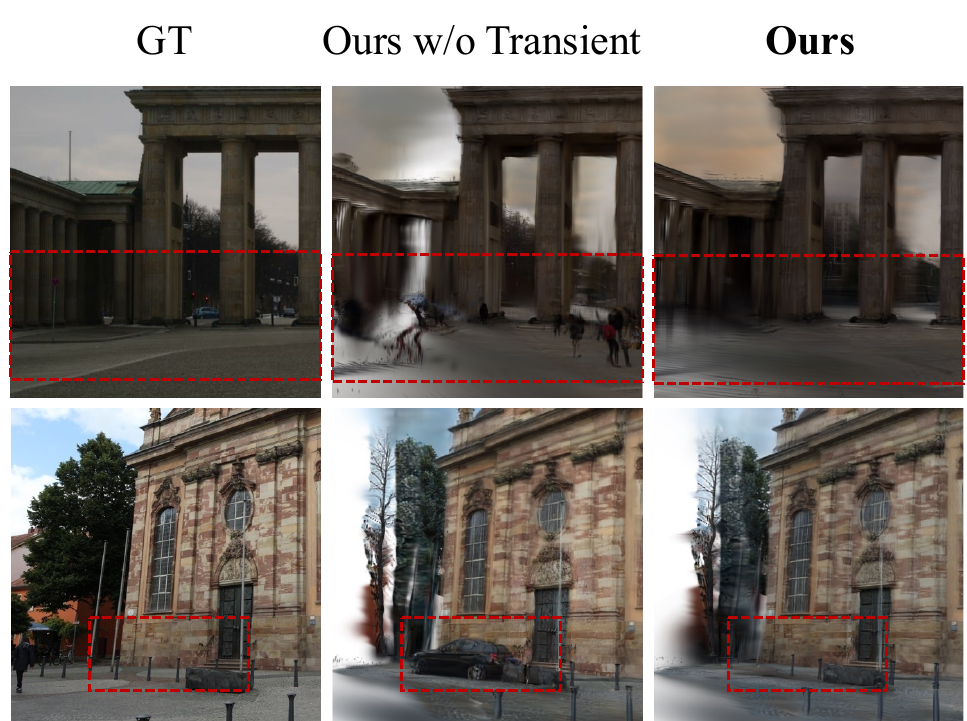}
    \caption{Effect of Adding Transient Objects.}
    \label{fig:ablation_transient}
\end{subfigure}\hfill
\begin{subfigure}{0.49\linewidth}
    \centering
    \includegraphics[width=\linewidth]{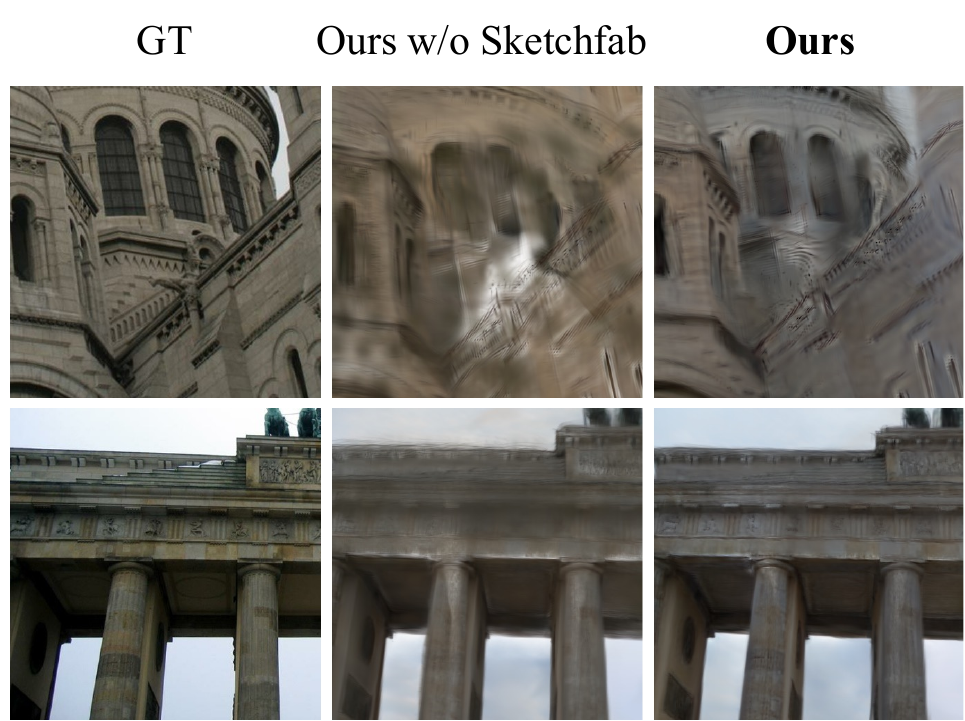}
    \caption{Effect of Sketchfab Assets.}
    \label{fig:ablation_asset}
\end{subfigure}

\caption{\textbf{Qualitative Ablation Study with 16 Context Views.} (a) Our full model enables transient object removal while the variant fails. (b) Our full model is more consistent on fine-grained geometries than the model trained without Sketchfab assets.}
\label{fig:ablation}
\vspace{-1em}
\end{figure}

In the data comparison, we also consider two variants: a model trained without transient objects added by Gemini~\cite{gemini} (c) and a model trained without Sketchfab assets (d).
As shown in the table, removing specific data components tends to degrade performance, demonstrating that data design plays a critical role. 
For example, excluding transient objects (c) leads to a noticeable performance drop compared to our full model.
This indicates that explicitly exposing the network to transient objects during training enhances its robustness, as highlighted by the red boxes in \Fref{fig:ablation_transient}.
Similarly, excluding assets sourced from Sketchfab (d) results in inferior performance.
This can be interpreted as evidence that incorporating diverse variations in shape, material, and appearance into training improves generalization and stability, as shown in \Fref{fig:ablation_asset}. 

Overall, our model consistently achieves robust and strong performance regardless of the number of context views. 
These results demonstrate that our improvements are attributed to both the fine-tuning strategy and our comprehensive data design, which provides essential shape diversity via Sketchfab and realistic transient distributions generated by Gemini.

\begin{table}[t]
    \centering 
    \begin{adjustbox}{width=1.0\linewidth}
    \begin{tabular}{lccccccccc} 
    \toprule
    \multirow{2}{*}{Method} & \multicolumn{3}{c}{4 Context Views}  & \multicolumn{3}{c}{16 Context Views} & \multicolumn{3}{c}{64 Context Views}\\
    \cmidrule(lr){2-4}\cmidrule(lr){5-7}  \cmidrule(lr){8-10}
    & PSNR$\uparrow$ &  SSIM$\uparrow$ & LPIPS$\downarrow$& PSNR$\uparrow$ &  SSIM$\uparrow$ & LPIPS$\downarrow$& PSNR$\uparrow$ &  SSIM$\uparrow$ & LPIPS$\downarrow$\\  
    \midrule
    \multicolumn{2}{l}{\textit{\textbf{Variants on Modeling}}} \\
    (a) AnySplat & 11.25 & 0.320 & 0.593 & 13.72 & 0.377 & 0.546 & 14.88 & 0.417 & 0.512\\
    (b) AnySplat w/ Fine-tuning  & 11.47 & 0.339 & 0.588 & 14.18 & 0.403 & 0.543 & 14.98 & 0.442 & 0.516\\
    \midrule
    \multicolumn{2}{l}{\textit{\textbf{Variants on Data}}}\\
    (c) Ours w/o Transient Objects & 12.66 & 0.366 & \textbf{0.556} & 15.57 & 0.432 & 0.514 & 16.08 & 0.451 & 0.487\\
    (d) Ours w/o Sketchfab Assets & 12.70 & \textbf{0.372} & 0.570 & 15.01 & 0.425 & 0.531 & 15.36 & 0.449 & 0.508\\
    \midrule
    (e) Ours & \textbf{13.04} & 0.370 & \textbf{0.556} & \textbf{15.87} & \textbf{0.435} & \textbf{0.506} & \textbf{16.29} & \textbf{0.458} & \textbf{0.477}\\
    \bottomrule
    \end{tabular}
    \end{adjustbox}
    \vspace{1.5mm}
  \caption{\textbf{Ablation Study on the Photo Tourism Dataset.} The best values are highlighted in \textbf{bold}. Our method achieves the best performance in most metrics.}
  \label{tb:ablation}
  \vspace{-1em}
\end{table}

\section{Limitations} 
\label{sec:limitation}
While \modelname demonstrates impressive results in unconstrained 3D reconstruction, it has several limitations inherent to its design.
First, our method does not explicitly model physical material parameters, making it difficult to capture highly complex specular effects and reflections.
Second, because the scene appearance is anchored to the reference image, recovering accurate illumination becomes challenging when large portions of this reference view are occluded by transient objects.
Third, due to the feed-forward architecture, our model inherently struggles to synthesize fine-grained, high-frequency details as accurately as optimization-based methods. 
Finally, the overall reconstruction quality is bounded by the representational capacity of the underlying feed-forward backbone. 
However, as feed-forward 3DGS architectures continue to evolve, our learning strategy and dataset can be seamlessly integrated to yield even higher-quality reconstructions in the future.

\section{Conclusion}
\label{sec:conclusion}
We presented \modelname, a feed-forward 3D Gaussian Splatting approach that scales to unconstrained photo collections, avoiding the expensive per-scene optimization of traditional 3DGS. 
Our work is motivated by the fact that existing training data with joint multi-view coverage, diverse illumination, and transient variations remains scarce, making it difficult to learn robust scene representations.
To bridge this gap, we introduced the \datasetname dataset, which contains \numscenes scenes captured under \numlights lighting conditions with transient objects, totaling \numimages images. 
Leveraging the \datasetname dataset, our new training strategy encourages appearance consistency across viewpoints and illuminations while suppressing transient content.
Our experiments show that \modelname outperforms existing pose-free feed-forward approaches and achieves results comparable to optimization-based methods that rely on camera calibration or point cloud initialization.

\section{Acknowledgements}
This work was partially financially supported by JST ASPIRE Program, Japan, Grant Number JPMJAP2303; JST ACT-X (JPMJAX25C5); JST SPRING, Grant Number JPMJSP2108; and JSPS KAKENHI Grant Number 26K21245.

\bibliographystyle{abbrv}
\bibliography{main}

@String(CVPR  = {IEEE Conf. Comput. Vis. Pattern Recog.})

@String(ICCV  = {Int. Conf. Comput. Vis.})

@String(ECCV  = {Eur. Conf. Comput. Vis.})

@String(NeurIPS = {Adv. Neural Inform. Process. Syst.})

@String(ICLR  = {Int. Conf. Learn. Represent.})

@String(TMLR  = {Trans. Mach. Learn Res.})

@String(TOG   = {ACM Trans. Graph.})

@String(TIP   = {IEEE Trans. Image Process.})

@String(CVPR  = {CVPR})

@String(ICCV  = {ICCV})

@String(ECCV  = {ECCV})

@String(NeurIPS = {NeurIPS})

@String(ICLR  = {ICLR})

@String(TMLR  = {TMLR})

@String(TOG   = {ACM TOG})

@String(TIP   = {IEEE TIP})

@inproceedings{kerbl3Dgaussians,
    author       = {Bernhard Kerbl and Georgios Kopanas and Thomas Leimkühler and George Drettakis},
    title        = {3D Gaussian Splatting for Real-Time Radiance Field Rendering},
    booktitle = {TOG},
    year      = {2023},
}

@inproceedings{nerf,
 title={NeRF: Representing Scenes as Neural Radiance Fields for View Synthesis},
 author={Ben Mildenhall and Pratul P. Srinivasan and Matthew Tancik and Jonathan T. Barron and Ravi Ramamoorthi and Ren Ng},
 year={2020},
 booktitle={ECCV},
}

@article{gemini,
      title={Gemini: A Family of Highly Capable Multimodal Models}, 
      author={Team, Gemini and Anil, Rohan and Borgeaud, Sebastian and Alayrac, Jean-Baptiste and Yu, Jiahui and Soricut, Radu and Schalkwyk, Johan and Dai, Andrew M and Hauth, Anja and Millican, Katie and others},
      year={2023},
      journal={arXiv preprint arXiv:2312.11805}
}

@inproceedings{martinbrualla2020nerfw,
  author = {Martin-Brualla, Ricardo
            and Radwan, Noha
            and Sajjadi, Mehdi S. M.
            and Barron, Jonathan T.
            and Dosovitskiy, Alexey
            and Duckworth, Daniel},
  title = {NeRF in the Wild: Neural Radiance Fields for
           Unconstrained Photo Collections},
  booktitle = {CVPR},
  year={2021}
}

@inproceedings{chen2022hallucinated,
  title={Hallucinated neural radiance fields in the wild},
  author={Chen, Xingyu and Zhang, Qi and Li, Xiaoyu and Chen, Yue and Feng, Ying and Wang, Xuan and Wang, Jue},
  booktitle={CVPR},
  year={2022}
}

@inproceedings{sun2022neural,
  title={Neural 3d reconstruction in the wild},
  author={Sun, Jiaming and Chen, Xi and Wang, Qianqian and Li, Zhengqi and Averbuch-Elor, Hadar and Zhou, Xiaowei and Snavely, Noah},
  booktitle={SIGGRAPH},
  year={2022}
}

@inproceedings{k-planes,
  title={K-planes: Explicit radiance fields in space, time, and appearance},
  author={Fridovich-Keil, Sara and Meanti, Giacomo and Warburg, Frederik Rahb{\ae}k and Recht, Benjamin and Kanazawa, Angjoo},
  booktitle={CVPR},
  year={2023}
}

@article{kassab2025refinedfields,
    title={RefinedFields: Radiance Fields Refinement for Planar Scene Representations},
    author={Karim Kassab and Antoine Schnepf and Jean-Yves Franceschi and Laurent Caraffa and Jeremie Mary and Val{\'e}rie Gouet-Brunet},
    journal={TMLR},
    year={2025}
}

@inproceedings{yang2023cross,
  title={Cross-Ray Neural Radiance Fields for Novel-view Synthesis from Unconstrained Image Collections},
  author={Yang, Yifan and Zhang, Shuhai and Huang, Zixiong and Zhang, Yubing and Tan, Mingkui},
  booktitle={ICCV},
  year={2023}
}

@inproceedings{swag,
  title={SWAG: Splatting in the Wild images with Appearance-conditioned Gaussians},
  author={Hiba Dahmani and Moussab Bennehar and Nathan Piasco and Luis Roldao and Dzmitry Tsishkou},
  booktitle={ECCV},
  year={2024}
}

@inproceedings{zhang2024GS-W,
  title={Gaussian in the wild: 3d gaussian splatting for unconstrained image collections},
  author={Zhang, Dongbin and Wang, Chuming and Wang, Weitao and Li, Peihao and Qin, Minghan and Wang, Haoqian},
  booktitle={ECCV},
  year={2024},
}

@inproceedings{wild-gs,
  title={Wild-GS: Real-Time Novel View Synthesis from Unconstrained Photo Collections},
  author={Jiacong Xu and Yiqun Mei and Vishal M. Patel},
  booktitle={NeurIPS},
  year={2024},
}

@inproceedings{kulhanek2024wildgaussians,
  title={{W}ild{G}aussians: {3D} Gaussian Splatting in the Wild},
  author={Kulhanek, Jonas and Peng, Songyou and Kukelova, Zuzana and Pollefeys, Marc and Sattler, Torsten},
  booktitle={NeurIPS},
  year={2024}
}

@inproceedings{
    li2025asymgs,
    title={Robust Neural Rendering in the Wild with Asymmetric Dual 3D Gaussian Splatting},
    author={Chengqi Li and Zhihao Shi and Yangdi Lu and Wenbo He and Xiangyu Xu},
    booktitle={NeurIPS},
    year={2025},
}

@inproceedings{
    wildsplatting,
    title={WildSplatting: Unposed Incremental 3D Gaussian Splatting Reconstruction in the Wild},
    author={Lifan Wu and Tianzhu Zhang},
    booktitle={VRCAI},
    year={2025},
}

@inproceedings{dust3r,
      title={DUSt3R: Geometric 3D Vision Made Easy}, 
      author={Shuzhe Wang and Vincent Leroy and Yohann Cabon and Boris Chidlovskii and Jerome Revaud},
      booktitle = {CVPR},
      year = {2024}
}

@article{jiang2025anysplat,
  title={Anysplat: Feed-forward 3d gaussian splatting from unconstrained views},
  author={Jiang, Lihan and Mao, Yucheng and Xu, Linning and Lu, Tao and Ren, Kerui and Jin, Yichen and Xu, Xudong and Yu, Mulin and Pang, Jiangmiao and Zhao, Feng and others},
  journal={TOG},
  year={2025},
}

@inproceedings{
    depthanything3,
    title={Depth Anything 3: Recovering the visual space from any views},
    author={Haotong Lin and Sili Chen and Jun Hao Liew and Donny Y. Chen and Zhenyu Li and Yang Zhao and Sida Peng and Hengkai Guo and Xiaowei Zhou and Guang Shi and Jiashi Feng and Bingyi Kang},
    booktitle={ICLR},
    year={2026},
}

@article{splatt3r,
  title={Splatt3R: Zero-shot Gaussian Splatting from Uncalibrated Image Pairs},
  author={Brandon Smart and Chuanxia Zheng and Iro Laina and Victor Adrian Prisacariu},
  journal={arXiv preprint arXiv:2408.13912},
  year={2024}
}

@inproceedings{
    flare,
    title={FLARE: Feed-forward Geometry, Appearance and Camera Estimation from Uncalibrated Sparse Views},
    author={Shangzhan Zhang and Jianyuan Wang and Yinghao Xu and Nan Xue and Christian Rupprecht and Xiaowei Zhou and Yujun Shen and Gordon Wetzstein},
    booktitle={CVPR},
    year={2025},
}

@inproceedings{
    yonosplat,
    title={YoNoSplat: You Only Need One Model for Feedforward 3D Gaussian Splatting},
    author={Botao Ye and Boqi Chen and Haofei Xu and Daniel Barath and Marc Pollefeys},
    booktitle={ICLR},
    year={2026},
}

@InProceedings{offthegrid,
  title={Off The Grid: Detection of Primitives for Feed-Forward 3D Gaussian Splatting},
  author={Moreau, Arthur and Shaw, Richard and Nazarczuk, Michal and Shin, Jisu and Tanay, Thomas and Zhang, Zhensong and Xu, Songcen and P{\'e}rez-Pellitero, Eduardo},
  booktitle={CVPR},
  year={2026}
}

@InProceedings{anchorsplat,
  title={AnchorSplat: Feed-Forward 3D Gaussian Splatting with 3D Geometric Priors}, 
  author={Xiaoxue Zhang and Xiaoxu Zheng and Yixuan Yin and Tiao Zhao and Kaihua Tang and Michael Bi Mi and Zhan Xu and Dave Zhenyu Chen},
  booktitle={CVPR},
  year={2026}
}

@inproceedings{longlrm,
  title={Long-LRM: Long-sequence Large Reconstruction Model for Wide-coverage Gaussian Splats},
  author={Chen Ziwen and Hao Tan and Kai Zhang and Sai Bi and Fujun Luan and Yicong Hong and Li Fuxin and Zexiang Xu},
  booktitle={ICCV},
  year={2025}
}

@article{phototourism,
  title={Photo tourism: exploring photo collections in 3d},
  author={Noah Snavely and Steven M Seitz and Richard Szeliski},
  journal={TOG},
  year={2006}
}

@InProceedings{rudnev2022nerfosr,
  title={NeRF for Outdoor Scene Relighting},
  author={Viktor Rudnev and Mohamed Elgharib and William Smith and Lingjie Liu and Vladislav Golyanik and Christian Theobalt},
  booktitle={ECCV},
  year={2022}
}

@InProceedings{lightcity,
  title={LightCity: An Urban Dataset for Outdoor Inverse Rendering and Reconstruction under Multi-illumination Conditions},
  author={Jingjing Wang and Qirui Hu and Chong Bao and Yuke Zhu and Hujun Bao and Zhaopeng Cui and Guofeng Zhang},
  booktitle={ICCV},
  year={2025}
}

@Misc{scenecity,
  title        = "{SceneCity}",
  author = {Arnaud Couturier},
  howpublished = "\url{https://www.cgchan.com/}"
}

@InProceedings{dtu,
  title={Large Scale Multi-view Stereopsis Evaluation},
  author={Rasmus Jensen and Anders Dahl and George Vogiatzis and Engin Tola and Henrik Aanæs},
  booktitle={CVPR},
  year={2014}
}

@InProceedings{Toschi_2023_CVPR,
  author    = {Toschi, Marco and De Matteo, Riccardo and Spezialetti, Riccardo and De Gregorio, Daniele and Di Stefano, Luigi and Salti, Samuele},
  title     = {ReLight My NeRF: A Dataset for Novel View Synthesis and Relighting of Real World Objects},
  booktitle = {CVPR},
  year      = {2023},
}

@InProceedings{openillumination,
  author    = {Isabella Liu and Linghao Chen and Ziyang Fu and Liwen Wu and Haian Jin and Zhong Li and Chin Ming Ryan Wong and Yi Xu and Ravi Ramamoorthi and Zexiang Xu and Hao Su},
  title     = {OpenIllumination: A Multi-Illumination Dataset for Inverse Rendering Evaluation on Real Objects},
  booktitle = {NeurIPS},
  year      = {2023},
}

@InProceedings{opensubstance,
  author    = {Fan Pei and Jinchen Bai and Xiang Feng and Zoubin Bi and Kun Zhou and Hongzhi Wu},
  title     = {OpenSubstance: A High-Quality Measured Dataset of Multi-View and -Lighting Images and Shapes},
  booktitle = {ICCV},
  year      = {2025},
}

@inproceedings{wang2025vggt,
  title={Vggt: Visual geometry grounded transformer},
  author={Wang, Jianyuan and Chen, Minghao and Karaev, Nikita and Vedaldi, Andrea and Rupprecht, Christian and Novotny, David},
  booktitle={CVPR},
  year={2025}
}

@article{ssim,
  title={Image quality assessment: from error visibility to structural similarity},
  author={Wang, Zhou and Bovik, Alan C and Sheikh, Hamid R and Simoncelli, Eero P.},
  journal={TIP},
  year={2004},
}

@inproceedings{lpips,
  title={The unreasonable effectiveness of deep features as a perceptual metric},
  author={Zhang, Richard and Isola, Phillip and Efros, Alexei A and Shechtman, Eli and Wang, Oliver},
  booktitle={CVPR},
  year={2018}
}

@inproceedings{schoenberger2016sfm,
    author={Johannes Lutz Sch\"{o}nberger and Jan-Michael Frahm},
    title={{Structure-from-Motion Revisited}},
    booktitle={CVPR},
    year={2016}
}

@inproceedings{co3dv2,
    author={Jeremy Reizenstein and Roman Shapovalov and Philipp Henzler and Luca Sbordone and Patrick Labatut and David Novotny},
    title={Common Objects in 3D: Large-Scale Learning and Evaluation of Real-life 3D Category Reconstruction},
    booktitle={ICCV},
    year={2021}
}

@inproceedings{blendedmvs,
    author={Yao Yao and Zixin Luo and Shiwei Li and Jingyang Zhang and Yufan Ren and Lei Zhou and Tian Fang and Long Quan},
    title={BlendedMVS: A Large-scale Dataset for Generalized Multi-view Stereo Networks},
    booktitle={CVPR},
    year={2020}
}

@inproceedings{dl3dv,
    author={Lu Ling and Yichen Sheng and Zhi Tu and Wentian Zhao and Cheng Xin and Kun Wan and Lantao Yu and Qianyu Guo and Zixun Yu and Yawen Lu and Xuanmao Li and Xingpeng Sun and Rohan Ashok and Aniruddha Mukherjee and Hao Kang and Xiangrui Kong and Gang Hua and Tianyi Zhang and Bedrich Benes and Aniket Bera},
    title={DL3DV-10K: A Large-Scale Scene Dataset for Deep Learning-based 3D Vision},
    booktitle={CVPR},
    year={2024}
}

@inproceedings{megadepth,
    author={Zhengqi Li and Noah Snavely},
    title={MegaDepth: Learning Single-View Depth Prediction from Internet Photos},
    booktitle={CVPR},
    year={2018}
}

@inproceedings{kubric,
    author={Klaus Greff and Francois Belletti and Lucas Beyer and Carl Doersch and Yilun Du and Daniel Duckworth and David J. Fleet and Dan Gnanapragasam and Florian Golemo and Charles Herrmann and Thomas Kipf and Abhijit Kundu and Dmitry Lagun and Issam Laradji and Hsueh-Ti (Derek)Liu and Henning Meyer and Yishu Miao and Derek Nowrouzezahrai and Cengiz Oztireli and Etienne Pot and Noha Radwan and Daniel Rebain and Sara Sabour and Mehdi S. M. Sajjadi and Matan Sela and Vincent Sitzmann and Austin Stone and Deqing Sun and Suhani Vora and Ziyu Wang and Tianhao Wu and Kwang Moo Yi and Fangcheng Zhong and Andrea Tagliasacchi},
    title={Kubric: A scalable dataset generator},
    booktitle={CVPR},
    year={2022}
}

@inproceedings{wildrgb,
  title={RGBD Objects in the Wild: Scaling Real-World 3D Object Learning from RGB-D Videos},
  author={Hongchi Xia and Yang Fu and Sifei Liu and Xiaolong Wang},
  booktitle={CVPR},
  year={2024}
}

@inproceedings{scannet,
    author={Angela Dai and Angel X. Chang and Manolis Savva and Maciej Halber and Thomas Funkhouser and Matthias Nießner},
    title={ScanNet: Richly-annotated 3D Reconstructions of Indoor Scenes},
    booktitle={CVPR},
    year={2017}
}

@inproceedings{hypersim,
    author={Mike Roberts and Jason Ramapuram and Anurag Ranjan and Atulit Kumar and Miguel Angel Bautista and Nathan Paczan and Russ Webb and Joshua M. Susskind},
    title={Hypersim: A Photorealistic Synthetic Dataset for Holistic Indoor Scene Understanding},
    booktitle={ICCV},
    year={2021}
}

@inproceedings{mapillary,
    author={Manuel Lopez Antequera and Pau Gargallo and Markus Hofinger and Samuel Rota Bulo and Yubin Kuang and Peter Kontschieder},
    title={Mapillary Planet-Scale Depth Dataset},
    booktitle={ECCV},
    year={2020}
}

@inproceedings{habitat,
    author={Andrew Szot and Alex Clegg and Eric Undersander and Erik Wijmans and Yili Zhao and John Turner and Noah Maestre and Mustafa Mukadam and Devendra Chaplot and Oleksandr Maksymets and Aaron Gokaslan and Vladimir Vondrus and Sameer Dharur and Franziska Meier and Wojciech Galuba and Angel Chang and Zsolt Kira and Vladlen Koltun and Jitendra Malik and Manolis Savva and Dhruv Batra},
    title={Habitat 2.0: Training Home Assistants to Rearrange their Habitat},
    booktitle={NeurIPS},
    year={2021}
}

@article{replica,
  title={The Replica Dataset: A Digital Replica of Indoor Spaces},
  author={Julian Straub and Thomas Whelan and Lingni Ma and Yufan Chen and Erik Wijmans and Simon Green and Jakob J. Engel and Raul Mur-Artal and Carl Ren and Shobhit Verma and Anton Clarkson and Mingfei Yan and Brian Budge and Yajie Yan and Xiaqing Pan and June Yon and Yuyang Zou and Kimberly Leon and Nigel Carter and Jesus Briales and Tyler Gillingham and Elias Mueggler and Luis Pesqueira and Manolis Savva and Dhruv Batra and Hauke M. Strasdat and Renzo De Nardi and Michael Goesele and Steven Lovegrove and Richard Newcombe},
  journal={arXiv preprint arXiv:1906.05797},
  year={2019}
}

@inproceedings{mvssynth,
    author={Po-Han Huang and Kevin Matzen and Johannes Kopf and Narendra Ahuja and Jia-Bin Huang},
    title={DeepMVS: Learning Multi-view Stereopsis},
    booktitle={CVPR},
    year={2018}
}

@inproceedings{pointodyssey,
    author={Yang Zheng and Adam W. Harley and Bokui Shen and Gordon Wetzstein and Leonidas J. Guibas},
    title={PointOdyssey: A Large-Scale Synthetic Dataset for Long-Term Point Tracking},
    booktitle={ICCV},
    year={2023}
}

@article{virtualkitti,
  title={Virtual KITTI 2},
  author={Yohann Cabon and Naila Murray and Martin Humenberger},
  journal={arXiv preprint arXiv:2001.10773},
  year={2020}
}

@inproceedings{aria,
    author={Xiaqing Pan and Nicholas Charron and Yongqian Yang and Scott Peters and Thomas Whelan and Chen Kong and Omkar Parkhi and Richard Newcombe and Carl Yuheng Ren},
    title={Aria Digital Twin: A New Benchmark Dataset for Egocentric 3D Machine Perception},
    booktitle={ICCV},
    year={2023}
}

@inproceedings{objaverse,
    author={Matt Deitke and Dustin Schwenk and Jordi Salvador and Luca Weihs and Oscar Michel and Eli VanderBilt and Ludwig Schmidt and Kiana Ehsani and Aniruddha Kembhavi and Ali Farhadi},
    title={Objaverse: A Universe of Annotated 3D Objects},
    booktitle={CVPR},
    year={2023}
}

\newpage
\appendix

\section{Implementation Details}

\paragraph{Scene and Camera Setup.}
During the manual selection of buildings for scene generation, we generally maintained a distance of at least 20 meters between the centers of adjacent scenes so that they are distinct scenes.
For the camera setup, the positions were uniformly sampled at a distance of 10 to 25 meters from the scene center, constrained within a 120-degree fan-shaped region.
The look-at point of each camera was offset from the scene center to introduce natural variation.
Specifically, the $x$ and $y$ coordinates of the look-at point were uniformly sampled from $[-2, 2]$ meters.
The $z$ coordinate was uniformly sampled from a manually selected range of $[1.5, 3]$, $[1.5, 5]$, $[1.5, 7]$, $[1.5, 10]$, or $[1.5, 12]$ meters, depending on the height of the target building.

\paragraph{Prompting for Adding Transient Objects.}
We use two types of prompts for Gemini (gemini-3-pro-image-preview), depending on the image content:

\noindent 1) If there are roads where people and cars can be placed, we use:

\noindent``Please add $t_p$, $t_v$, $t_s$, $t_b$, 
while maintaining the geometry and lightness/darkness.''

\noindent 2) If not, we use:

\noindent``Please add $\tilde{t}_b$, 
while maintaining the geometry and lightness/darkness.''

\noindent where the words $t_p$, $t_v$, $t_s$, $t_b$, and $\tilde{t}_b$ are sampled from $T_p$, $T_v$, $T_s$, $T_b$, and $\widetilde{T}_b$ as in \Tref{tb:prompt}, respectively.
We classify each image into the two types above,
using another variant of Gemini (gemini-3-flash-preview) with a predefined prompt $q$, as shown in \Tref{tb:prompt}.

\begin{figure*}[t]
\centering
\includegraphics[width=\linewidth]{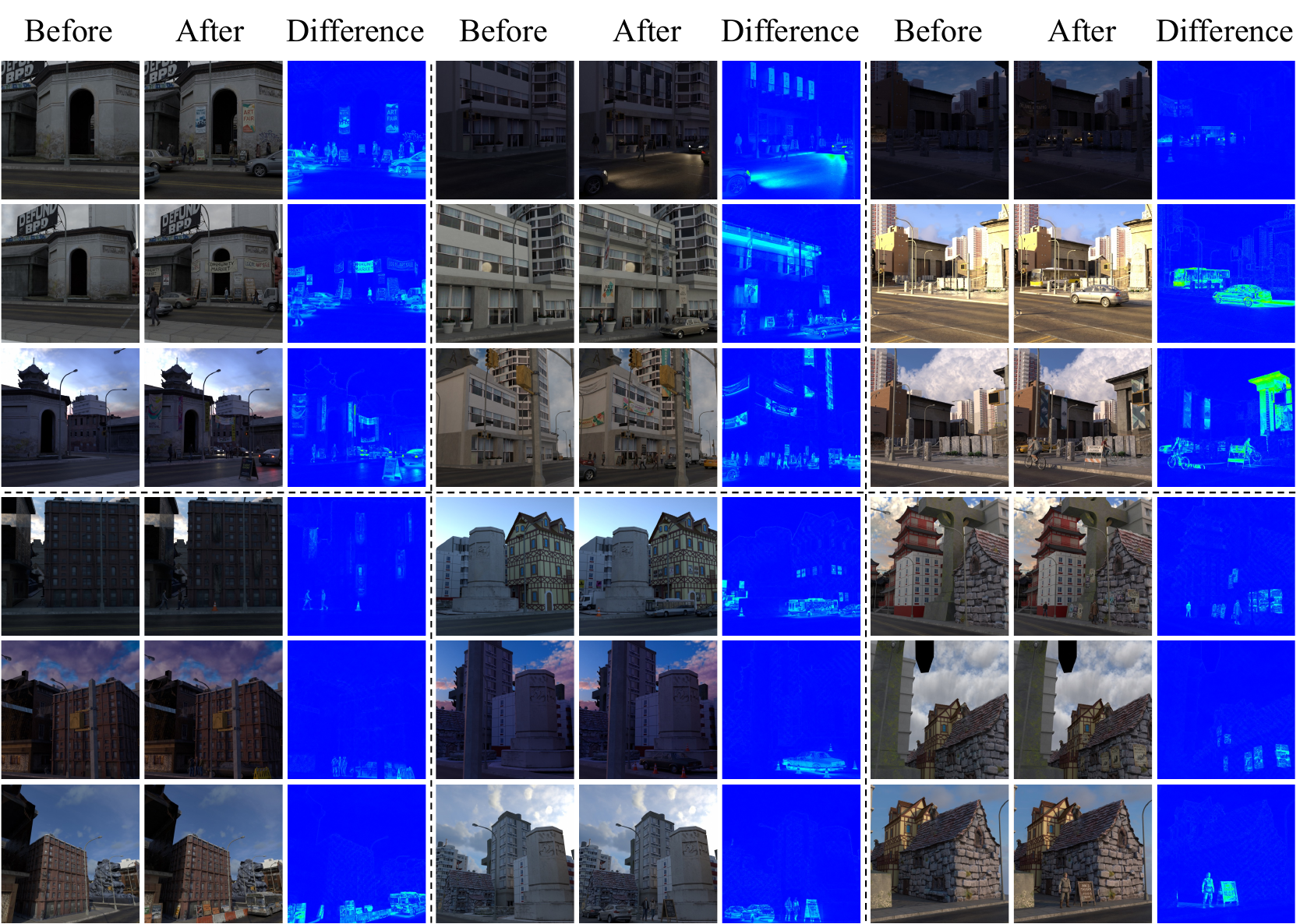}
\caption{\textbf{Qualitative Validation of Adding Transient Objects.} We show randomly selected examples from the \datasetname dataset before and after adding transient objects via a text-driven image editing model~\cite{gemini}, alongside their corresponding pixel-wise difference heatmaps.}
\label{fig:diff}
\end{figure*}

\paragraph{Validity of Adding Transient Objects.}
To validate our 2D augmentation strategy, we qualitatively analyze the effect of adding transient objects.
\Fref{fig:diff} illustrates randomly selected pairs of images before and after the augmentation, alongside their corresponding pixel-wise difference heatmaps.
As observed in the difference maps, the modifications introduced by a text-driven image editing model~\cite{gemini} are highly localized.
Crucially, while local illumination variations naturally occur around the new objects, the global lighting conditions and the underlying static geometry of the background are consistently preserved without destructive artifacts.
This qualitative evidence confirms that our augmentation successfully introduces complex and realistic objects without corrupting multi-view consistency of the static background.

\paragraph{Training.}
To robustify the model against unconstrained photo collections, we apply the following spatial and appearance augmentations. 
Following AnySplat~\cite{jiang2025anysplat}, the maximum input resolution is scaled to 448 pixels on the longer edge, with the aspect ratio randomized within $[0.5, 1.0]$.
For each image independently, we apply random anisotropic scaling (scale factors $\in [0.9, 1.1]$, $p=0.1$) and random resizing (scale factors $\in [0.7, 1.2]$, $p=0.3$). 
If the resulting image is smaller than the target resolution, it is upscaled. 
This is followed by either a random crop ($p=0.3$) or a center crop ($p=0.7$).
Subsequently, the following appearance augmentations are applied independently to each image: 
JPEG compression (quality $\in [50, 100]$, $p=0.3$), Gaussian noise ($\sigma \in [0, 0.03]$, $p=0.3$), and color jitter (brightness, contrast, saturation, and hue factors up to 0.1, $p=0.3$). 
Alternatively, with $p=0.1$, the image is converted to grayscale instead of applying color jitter. 
Importantly, whenever color jitter or grayscale conversion is applied to the reference view, the exact same transformation is applied to all target images to ensure photometric consistency during supervision.

\paragraph{Evaluation.}
To quantitatively evaluate the reconstruction quality, we employ three widely used metrics: Peak Signal-to-Noise Ratio (PSNR), Structural Similarity Index Measure (SSIM)~\cite{ssim}, and Learned Perceptual Image Patch Similarity (LPIPS)~\cite{lpips}.

PSNR measures the pixel-wise reconstruction fidelity between the ground truth image $I$ and the rendered image $\hat{I}$. It is defined as
\begin{equation}
\mathrm{PSNR} = 10 \log_{10} \left( \frac{I_\mathrm{MAX}^2}{\mathrm{MSE}(I, \hat{I})} \right),    
\end{equation}
where $I_\mathrm{MAX}$ denotes the maximum possible pixel value and $\mathrm{MSE}(I, \hat{I})$ is the Mean Squared Error (MSE), computed as
$\mathrm{MSE}(I, \hat{I}) = \frac{1}{N}\sum_{i=1}^{N}(I_i - \hat{I}_i)^2$. Here, $N$ represents the total number of pixels, and $I_i, \hat{I}_i$ denote the color values of the $i$-th pixel in the images, respectively.

SSIM evaluates the perceptual similarity by comparing luminance, contrast, and structural information between two images. It is defined as
\begin{equation}
\mathrm{SSIM}(I,\hat{I}) =
\frac{(2\mu_I\mu_{\hat{I}} + C_1)(2\sigma_{I\hat{I}} + C_2)}
{(\mu_I^2 + \mu_{\hat{I}}^2 + C_1)(\sigma_I^2 + \sigma_{\hat{I}}^2 + C_2)},
\end{equation}
where $\mu$, $\sigma^2$, and $\sigma_{I\hat{I}}$ denote the mean, variance, and covariance of the images, respectively, and $C_1$ and $C_2$ are stabilization constants.

LPIPS measures perceptual similarity in a deep feature space. It computes the distance between normalized deep features:
\begin{equation}
\mathrm{LPIPS}(I,\hat{I}) =
\sum_l \frac{1}{H_l W_l}
\sum_{h,w} | w_l \odot (\phi_l(I)_{hw} - \phi_l(\hat{I})_{hw}) |_2^2 ,
\end{equation}
where $\phi_l$, $w_l$, $H_l$, $W_l$ denote the feature map from layer $l$ of a pretrained network, learned channel-wise weights, the height of $\phi_l$, and the width of $\phi_l$, respectively.

\begin{table}[t]
\centering
\begin{adjustbox}{width=0.85\linewidth}
\begin{tabular}{l|p{10cm}}
\toprule
$T_p$ & \{a person,
        two pedestrians,
        several pedestrians,
        a small group of people,
        a couple of cyclists,
        a delivery,
        a delivery worker and a few pedestrians\}\\
$T_v$ & \{a car,
        a few cars,
        a taxi,
        a taxi and a car,
        a van,
        a van and a car,
        a parked car,
        a couple of parked cars,
        a bus,
        a bus and a car\}\\
$T_s$ &\{a traffic cone,
        traffic cones,
        a temporary barrier,
        temporary barriers,
        a portable sign stand,
        a construction warning sign,
        construction warning signs\} \\
$T_b$ & \{a fabric banner on a building,
        fabric banners on some buildings,
        a removable signboard on a storefront,
        removable signboards on storefronts,
        an event poster on a wall,
        event posters on walls,
        some cracks in buildings\}\\
$\widetilde{T}_b$& \{a fabric banner on a building,
        fabric banners on some buildings,
        an event poster on a wall,
        event posters on walls,
        some cracks in buildings\}\\
$q$& \textit{As data augmentation of the city image data, 
        I want to add people and cars using image editing. 
        However, I first want to determine whether there are locations (i.e., streets) where people and cars could realistically be present for adding them. 
        Do you think this image is suitable in this situation? 
        Just answer by '1' (Yes) or '0' (No)}\\
\bottomrule
\end{tabular}
\end{adjustbox}
\vspace{1.5mm}
\caption{\textbf{Prompt templates.}}
\label{tb:prompt}
\end{table}

\noindent\textbf{Evaluation Protocol.} 
In evaluation, feed-forward methods take the $N$ context images together with the reference half-image as input.
For \modelname, this reference provides appearance conditioning, whereas for other feed-forward baselines it simply acts as an additional input view.
Optimization-based methods are trained using the same $N$ context images, and methods that support appearance adaptation~\cite{martinbrualla2020nerfw,zhang2024GS-W,kulhanek2024wildgaussians,li2025asymgs} further optimize their appearance embeddings using the reference half-image.
To match the input resolution expected by feed-forward models~\cite{jiang2025anysplat}, all images are resized and cropped to $448\times448$ with bottom alignment to ensure transient objects are retained.
Additionally, camera-free feed-forward approaches determine their camera poses following the same procedure as AnySplat~\cite{jiang2025anysplat}.

\section{Additional Experiments}
\subsection{Additional Qualitative Comparison on the Photo Tourism Dataset.}
\begin{figure*}[t]
\centering
\includegraphics[width=\linewidth]{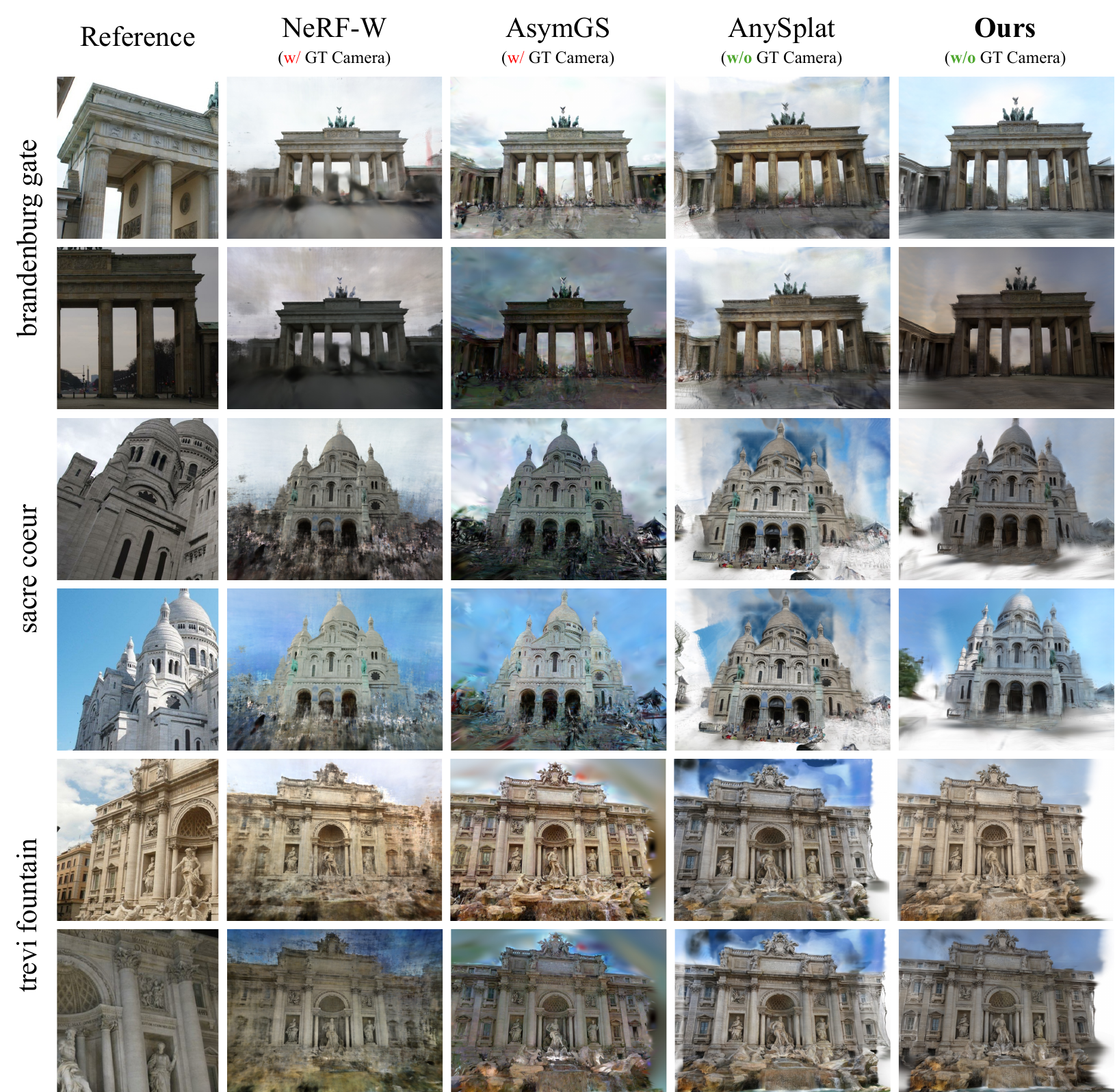}
\caption{\textbf{Additional Qualitative Comparison on the Photo Tourism Dataset.}}
\label{fig:nerfw-more}
\vspace{-5mm}
\end{figure*}

We report the evaluations of the models on the Photo Tourism dataset~\cite{phototourism} in the main paper.
However, under standard evaluation protocols, the test images predominantly consist of close-up views without transient objects, precluding quantitative evaluation on global views that capture the entire scene.
Furthermore, for camera-free feed-forward methods, quantitative metrics are highly sensitive to inaccuracies in camera pose estimation.
Therefore, we present qualitative comparisons of the buildings from distant, global views in \Fref{fig:nerfw-more}.
As observed, compared to AnySplat~\cite{jiang2025anysplat}, \modelname effectively adapts to varying illumination guided by the provided reference images.
Furthermore, in contrast to optimization-based methods such as NeRF-W~\cite{martinbrualla2020nerfw} and AsymGS~\cite{li2025asymgs}, our approach suppresses artifacts caused by transient objects and variable illumination.

\subsection{Evaluation on the NeRF-OSR Dataset}
We also evaluate on the NeRF-OSR dataset~\cite{rudnev2022nerfosr} using four commonly used scenes: stjohann, lwp, st, and europa. 
We construct the test sets in the same manner as the Photo Tourism dataset~\cite{phototourism} and report the per-scene metrics with 16 context views.

\noindent\textbf{Quantitative Comparison.}
A quantitative comparison on the NeRF-OSR dataset is reported in \Tref{tb:nerfosr}. 
Our proposed method, \modelname, consistently outperforms the feed-forward baseline AnySplat. 
Even compared with optimization-based approaches, \modelname achieves competitive results, particularly on perceptual metrics such as LPIPS, despite not requiring camera parameters.

\begin{figure*}[t]
\centering
\includegraphics[width=\linewidth]{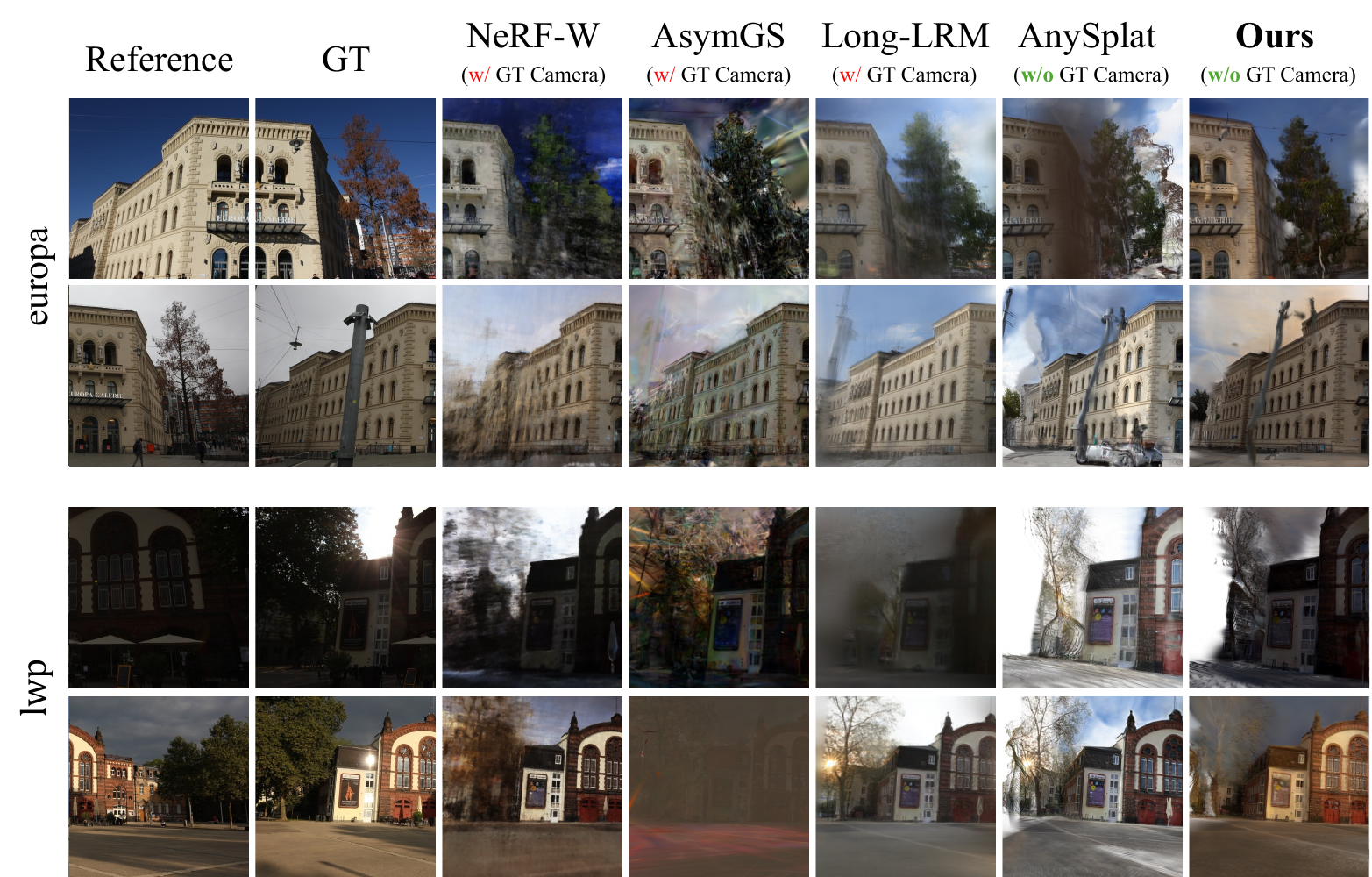}
\caption{\textbf{Qualitative Comparison on the NeRF-OSR Dataset.}}
\label{fig:nerfosr}
\end{figure*}

\noindent\textbf{Qualitative Comparison.}
A qualitative comparison on the NeRF-OSR dataset is shown in \Fref{fig:nerfosr}.
While existing optimization-based approaches exhibit artifacts in sparse-view settings, our method mitigates artifacts caused by the sky and transient objects.
Furthermore, compared to AnySplat~\cite{jiang2025anysplat}, our method is capable of accurately modeling substantial illumination changes, faithfully reproducing lighting conditions with dark-to-bright contrasts.

\begin{table}[t]
    \centering 
    \begin{adjustbox}{width=1.0\linewidth}
    \begin{tabular}{lcccccccccccccc} 
    \toprule
    \multirow{2}{*}{Method} & \multirow{2}{*}{Unknown}& \multirow{2}{*}{Unknown} &\multicolumn{3}{c}{europa} & \multicolumn{3}{c}{lwp} & \multicolumn{3}{c}{st} & \multicolumn{3}{c}{stjohann}  \\
    \cmidrule(lr){4-6}\cmidrule(lr){7-9}  \cmidrule(lr){10-12} \cmidrule(lr){13-15}
    &Camera & Point Cloud& PSNR$\uparrow$ &  SSIM$\uparrow$ & LPIPS$\downarrow$& PSNR$\uparrow$ &  SSIM$\uparrow$ & LPIPS$\downarrow$& PSNR$\uparrow$ &  SSIM$\uparrow$ & LPIPS$\downarrow$& PSNR$\uparrow$ &  SSIM$\uparrow$ & LPIPS$\downarrow$\\  
    \midrule
    \multicolumn{2}{l}{\textit{Optimization-based}} \\
    NeRF-W~\cite{martinbrualla2020nerfw} &{\color{red}\ding{55}}&{\color{red}\ding{55}} & \textbf{14.26} & \textbf{0.474} & 0.613 & \textbf{13.90} & \textbf{0.442} & \textbf{0.596} & \textbf{15.25} & \textbf{0.487} & 0.606 & \textbf{14.23} & \textbf{0.530} & 0.558\\
    3DGS~\cite{kerbl3Dgaussians} &{\color{red}\ding{55}}&{\color{red}\ding{55}} & 11.05 & 0.284 & 0.631 & 9.86 & 0.263 & 0.632 & 10.34 & 0.282 & 0.625 & 10.16 & 0.326 & 0.631\\
    GS-W~\cite{zhang2024GS-W} &{\color{red}\ding{55}}&{\color{red}\ding{55}} & 11.65 & 0.356 & \textbf{0.596} & 11.38 & 0.319 & 0.622 & 12.56 & 0.365 & 0.627 & 10.93 & 0.408 & \textbf{0.548}\\
    WildGaussians~\cite{kulhanek2024wildgaussians} &{\color{red}\ding{55}}&{\color{red}\ding{55}} & 11.38 & 0.354 & 0.640 & 11.78 & 0.323 & 0.645 & 12.70 & 0.357 & 0.632 & 11.28 & 0.370 & 0.601  \\
    AsymGS~\cite{li2025asymgs} &{\color{red}\ding{55}}&{\color{red}\ding{55}} & 12.75 & 0.409 & 0.616 & 12.35 & 0.350 & 0.635 & 13.81 & 0.389 & \textbf{0.601} & 13.56 & 0.491 & 0.572\\
    \midrule
    \multicolumn{2}{l}{\textit{Camera-known Feed-forward}}\\
    Long-LRM~\cite{longlrm} & {\color{red}\ding{55}}&{\color{Green}\ding{51}} & 13.23 & 0.470 & 0.544 & 11.86 & 0.373 & 0.609 & 12.86 & 0.434 & 0.570 & 13.84 & 0.518 & 0.542 \\
    \midrule
    \multicolumn{2}{l}{\textit{Camera-free Feed-forward}}\\
    AnySplat~\cite{jiang2025anysplat} &{\color{Green}\ding{51}} &{\color{Green}\ding{51}} & 11.18 & 0.330 & 0.555 & 9.47 & 0.234 & 0.608 & 10.13 & 0.272 & 0.616 & 10.70 & 0.356 & 0.583\\
    YoNoSplat~\cite{yonosplat} & {\color{Green}\ding{51}}&{\color{Green}\ding{51}} & 12.33 & 0.371 & 0.637 & 11.15 & 0.290 & 0.703 & 12.11 & \textbf{0.368} & 0.675 & 12.09 & 0.395 & 0.647\\
    DA3~\cite{depthanything3} & {\color{Green}\ding{51}}&{\color{Green}\ding{51}} & 12.67 & 0.378 & 0.589 & \textbf{11.47} & \textbf{0.307} & 0.638 & 12.24 & 0.352 & 0.606 & 11.84 & 0.365 & 0.604\\
    \modelname (Ours) &{\color{Green}\ding{51}} &{\color{Green}\ding{51}} & \textbf{13.27} & \textbf{0.388} & \textbf{0.535} & 10.68 & 0.278 & \textbf{0.598} & \textbf{12.67} & 0.356 & \textbf{0.584} & \textbf{12.22} & \textbf{0.401} & \textbf{0.547} \\
    \bottomrule
    \end{tabular}
    \end{adjustbox}
    \vspace{1.5mm}
  \caption{\textbf{Comparison with Previous Methods on the NeRF-OSR Dataset.} The best results for both optimization-based and camera-free feed-forward models are highlighted in \textbf{bold}.}
  \label{tb:nerfosr}
\end{table}

\subsection{Comparison with Other Datasets for Training.}
To further demonstrate the advantages of our \datasetname dataset, we trained our model on alternative datasets, including DTU~\cite{dtu} and LightCity~\cite{lightcity}, following the same protocol as the main paper. 
LightCity was not originally designed for training, as it contains only 5 scenes. 
Furthermore, since its depth maps are not provided, we generated pseudo ground truth depth maps from the ground truth camera parameters and images using COLMAP~\cite{schoenberger2016sfm}. 
These datasets inherently limit model generalization: first, both lack transient objects; second, LightCity provides insufficient scene diversity; finally, DTU is object-centric, lacking scene complexity and lighting variations.
As shown in \Tref{tb:dataset_ablation}, models trained on DTU or LightCity exhibit degraded performance, likely due to overfitting to their limited scene diversity. 
In contrast, the model trained on our \datasetname dataset achieves the highest performance across all metrics.
This clearly demonstrates the effectiveness of our proposed dataset for in-the-wild reconstruction.

\begin{table}[t]
    \centering 
    \begin{adjustbox}{width=1.0\linewidth}
    \begin{tabular}{lccccccccc} 
    \toprule
    \multirow{2}{*}{Training Dataset} & \multicolumn{3}{c}{4 Context Views}  & \multicolumn{3}{c}{16 Context Views} & \multicolumn{3}{c}{64 Context Views}\\
    \cmidrule(lr){2-4}\cmidrule(lr){5-7}  \cmidrule(lr){8-10}
    & PSNR$\uparrow$ &  SSIM$\uparrow$ & LPIPS$\downarrow$& PSNR$\uparrow$ &  SSIM$\uparrow$ & LPIPS$\downarrow$& PSNR$\uparrow$ &  SSIM$\uparrow$ & LPIPS$\downarrow$\\  
    \midrule
    \textit{None}  & 11.25 & 0.320 & 0.593 & 13.72 & 0.377 & 0.546 & 14.88 & 0.417 & 0.512\\
    \midrule
    DTU~\cite{dtu} & 11.29 & 0.299 & 0.629 & 12.79 & 0.352 & 0.599 & 13.51 & 0.391 & 0.585\\
    LightCity~\cite{lightcity} & 9.69 & 0.288 & 0.667 & 11.17 & 0.309 & 0.676 & 12.33 & 0.340 & 0.690\\
    \datasetname (Ours) & \textbf{13.04} & \textbf{0.370} & \textbf{0.556} & \textbf{15.87} & \textbf{0.435} & \textbf{0.506} & \textbf{16.29} & \textbf{0.458} & \textbf{0.477}\\
    \bottomrule
    \end{tabular}
    \end{adjustbox}
    \vspace{1mm}
  \caption{\textbf{Training Dataset Comparison on the Photo Tourism Dataset.} The best values are highlighted in \textbf{bold}. ``\textit{None}'' refers to AnySplat~\cite{jiang2025anysplat} without additional training.}
  \label{tb:dataset_ablation}
\end{table}

\subsection{Ablation Study on Freezing Pretrained Modules}
As mentioned in the main paper, we freeze the depth and camera heads during training. 
To validate this design choice, we evaluate a variant where the depth and camera heads are jointly updated during training.
As shown in \Tref{tb:freeze_ablation}, unfreezing these pretrained heads does not yield meaningful performance improvements. 
This indicates that the pretrained depth and camera heads already possess sufficient generalization capabilities, and updating them provides no additional benefit.

\begin{table}[ht]
    \centering 
    \begin{adjustbox}{width=1.0\linewidth}
    \begin{tabular}{lccccccccc} 
    \toprule
    \multirow{2}{*}{Method} & \multicolumn{3}{c}{4 Context Views}  & \multicolumn{3}{c}{16 Context Views} & \multicolumn{3}{c}{64 Context Views}\\
    \cmidrule(lr){2-4}\cmidrule(lr){5-7}  \cmidrule(lr){8-10}
    & PSNR$\uparrow$ &  SSIM$\uparrow$ & LPIPS$\downarrow$& PSNR$\uparrow$ &  SSIM$\uparrow$ & LPIPS$\downarrow$& PSNR$\uparrow$ &  SSIM$\uparrow$ & LPIPS$\downarrow$\\  
    \midrule
    Unfrozen & 12.95 & 0.367 & 0.557 & 15.77 & \textbf{0.438} & 0.508 & \textbf{16.30} & 0.453 & 0.487\\ 
    Ours (Frozen) & \textbf{13.04} & \textbf{0.370} & \textbf{0.556} & \textbf{15.87} & 0.435 & \textbf{0.506} & 16.29 & \textbf{0.458} & \textbf{0.477}\\
    \bottomrule
    \end{tabular}
    \end{adjustbox}
    \vspace{1mm}
  \caption{\textbf{Ablation Study on the Photo Tourism Dataset.} The best values are highlighted in \textbf{bold}. ``Unfrozen'' refers to training the depth and camera heads.}
  \label{tb:freeze_ablation}
\end{table}


\end{document}